\documentclass[runningheads]{llncs}

\usepackage{graphicx}
\usepackage{comment}
\usepackage{amsmath,amssymb}
\usepackage{color}
\usepackage{url}
\usepackage{hyperref}
\usepackage[export]{adjustbox}
\usepackage{lineno}

\newcommand{\pixel}[1] {\mathcal{#1}}
\newcommand{\parameters}[1] {\kappa_{\pixel{#1}}}
\newcommand{\sums}[1] {\tilde{#1}}

\titlerunning{A Framework for Noise Model-Aware Random Walker Image Segmentation}

\author{Dominik Drees\inst{1}$^{\ast}$ \and
Florian Eilers\inst{1}$^{\ast}$ \and
Ang Bian\inst{2} \and
Xiaoyi Jiang\inst{1}}

\begin{document}

\title{A Bhattacharyya Coefficient-Based Framework for Noise Model-Aware Random Walker Image Segmentation}

\authorrunning{D. Drees, F. Eilers, A. Bian and X. Jiang}
\institute{University of Münster, Münster, Germany \and
College of Computer Science, Sichuan University, Chengdu, China\\
$^{\ast}$ These authors contributed equally to this work.}

\maketitle              

\begin{abstract}
One well established method of interactive image segmentation is the random walker algorithm. Considerable research on this family of segmentation methods has been continuously conducted in recent years with numerous applications. These methods are common in using a simple Gaussian weight function which depends on a parameter that strongly influences the segmentation performance. In this work we propose a general framework of deriving weight functions based on probabilistic modeling. This framework can be concretized to cope with virtually any well-defined noise model. It eliminates the critical parameter and thus avoids time-consuming parameter search. We derive the specific weight functions for common noise types and show their superior performance on synthetic data as well as different biomedical image data (MRI images from the NYU fastMRI dataset, larvae images acquired with the FIM technique). Our framework can also be used in multiple other applications, e.g., the graph cut algorithm and its extensions.
\end{abstract}

\section{Introduction}
Interactive image segmentation aims to segment the objects of interest with minimal user input. Despite the remarkable performance achieved by deep learning techniques, interactive segmentation approaches remain attractive.
Supervised learning methods require accurate ground truth pixel-level training data, which is often expensive and tedious to produce -- a problem which interactive image segmentation algorithms (as sophisticated labeling tools) help to mitigate.
Supervised learning methods also tend to perform poorly on unseen object classes.
Furthermore, in application fields like biomedicine one may also be confronted with the problem of prohibitively small data sets.
Therefore, considerable research on interactive segmentation has been continuously conducted in recent years \cite{Li2018,Wang2021,Xu2021,Zhang2020}.

Random walker is one of the most popular interactive segmentation families~\cite{wang_rw_review} since its inception in Grady's seminal work~\cite{grady2006random}.
This method models a 2D or higher dimensional image as a graph and achieves a multi-region segmentation by calculating probabilities that each pixel is connected to the user provided class-specific seed pixels (seeds).
Considerable recent developments have been reported for this family of segmentation methods \cite{DBLP:journals/tgrs/CuiXMRM18,drees2021hierarchical,Ham2013,DBLP:journals/tip/KangZM20,DBLP:journals/tip/ShenDWL14,improvingrw2022wang} and numerous applications of random walker segmentation can be found in the literature~\cite{cheng2020organ,gerlein2019cygnss,mathewlynn2019volume,shi2016many}.
In particular, it plays an important role in biomedical imaging. For instance, the recent platform Biomedisa for biomedical image segmentation \cite{loesel2020biomedisa} is fully based on random walker.

Besides user provided seeds, random walker also requires a well-defined weight function for mapping the image information to a graph.
Grady \cite{grady2006random} used the simple Gaussian weight function, chosen for empirical reasons, which has become common practice and has also been adopted by the recent developments of random walker segmentation \cite{DBLP:journals/tgrs/CuiXMRM18,drees2021hierarchical,Ham2013,DBLP:journals/tip/KangZM20,DBLP:journals/tip/ShenDWL14,improvingrw2022wang}.
However, Bian et al.~\cite{ang2016statistical,ang2016ttest} showed that the optimal choice for its configurable parameter $\beta$ highly influences the segmentation performance and depends on the image conditions.
The optimal value varies even within rather homogeneous datasets and different regions of the same image.
Instead of the simple definition, we propose in this work a general framework of deriving weight functions based on probabilistic modeling.
This framework can be concretized to cope with virtually any well-defined noise model.
This approach avoids the critical parameter.
While previous methods on alternative weight functions work for specific noise models (additive Gaussian noise~\cite{ang2016statistical} and multiplicative speckle noise~\cite{ang2016ttest}), solutions for other noise models and in particular multi-channel images and noise are yet missing.
Our work extends the application spectrum of adaptive random walker methods to many domains (e.g., biomedical imaging, remote sensing), where imaging modalities with different noise models are of high importance.

In summary, with this work we \textit{contribute} the following:
1) A general framework for deriving pixel similarity measures for a given noise model and its utilization as a weight function (\autoref{sec:framework}).
2) Derivations of weight functions for specific noise models using the framework, in particular for Poisson and multivariate Gaussian noise (\autoref{sec:noise_models}).
3) Demonstration of on-par or superior performance compared to state of the art (\autoref{sec:results}).
4) An implementation of the presented work as an easy-to-use and open source python package available online\footnote{\url{https://zivgitlab.uni-muenster.de/ag-pria/rw-noise-model}}.

\section{Related Work}
\label{sec:relaetdwork}

There is very little work on weight function for random walker image segmentation.
Even most of the recent developments are based on the standard definition. The variant in \cite{sinop2007seeded} includes a spatial term (difference of two neighbors) into the weight function that has a minor influence.
Freedman \cite{freedman2012improved} proposes a weight function based on per-class density in LUV color space estimated from the colors of seed pixels.
As a consequence, semantically different regions in the image are expected to have distinct colors, which is likely not compatible with many (single channel) biomedical images.
Cerrone et al.~\cite{cerrone2019endtoend} propose a random walker based method with end-to-end learned weight function.
While very noteworthy and interesting theoretically, the need for labeled training data drastically reduces the usefulness of this method in practice as this is in direct conflict with the use as a labeling tool itself.

There are two major works we consider as direct prior work for noise model incorporation into random walker weight functions \cite{ang2016statistical,ang2016ttest}.
In \cite{ang2016statistical} additive Gaussian noise with constant global variance is assumed and the PDF of the estimated local means' difference is applied as an adaptive weight function. In \cite{ang2016ttest}, the signal-dependent local Gaussian model with variable regional variances is assumed for the multiplicative speckle noise (with additive Gaussian noise and Loupas noise as special cases). A statistical T-test based weight function is proposed.

Noise models have also been studied for image segmentation in other contexts. In \cite{chesnaud1999statistical,martin2004influence} it is shown that the noise type has an impact in active contours based image segmentation. The authors incorporate knowledge of the underlying noise model in an external energy to improve the results. A number of variational approaches have utilized knowledge about the underlying noise. This is done either by specifically designed data fidelity terms (e.g., for additive and multiplicative noise \cite{ali2018image}, Poisson noise, additive Gaussian noise and multiplicative speckle noise \cite{chen2013variational}) or variational frameworks \cite{sawatzky2013variational,TENBRINCK2015arbnoise} with concretization to particular noise models. Our work follows the general strategy of the latter approach and develops such a general framework for random walker segmentation.

\section{Framework}
\label{sec:framework}
As discussed in the introduction, random walker segmentation as introduced by Grady~\cite{grady2006random} is highly dependent on the choice of the parameter $\beta$.
In this section we present the general idea of the framework for random walker weight functions (independent of a concrete noise model) that is independent of such parameter usage.
After a brief introduction to random walker segmentation this includes the formulation of the weight function and a model for sampling concrete image pixels from the neighborhood of an edge that will be used to define the weight.

\subsection{Random Walker Segmentation}
\label{subsec:random_walker}
An image is defined as an undirected graph \(G(V,E)\), where \(V\) is the set of nodes corresponding to pixels and \(E\) is the set of weighted edges connecting adjacent nodes.
The random walker algorithm assigns each unmarked node a per-class probability, corresponding to the probability that a random walker starting in a marked seed reaches this node first.
The class with the highest probability is assigned to the node.
By partitioning the nodes into \(V_M\) (marked seed nodes) and \(V_U\) (unmarked nodes), the probabilities $P(V_U)$ for one class can be solved as a combinatorial Dirichlet problem with boundary conditions:
\begin{equation}
   \label{eq:linear_system}
P(V_U) = -(L_U)^{-1}B^T P(V_M); \quad \makebox{with components from }
L =
\begin{bmatrix}L_M & B\\B^T & L_U \end{bmatrix}
\end{equation}
where \(L\) is the Laplacian matrix with $L_{\pixel{X}\pixel{Y}}, \ \pixel{X}\not=\pixel{Y}$, equal to weight $-w_{\pixel{X}\pixel{Y}}$ for each edge, $\sum_{\pixel{Y}}{w_{\pixel{X}\pixel{Y}}}$ for diagonal elements $L_{\pixel{X}\pixel{X}}$, and 0 otherwise. A Gaussian weight function (with $x$/$y$ being the intensity in $\pixel{X}/\pixel{Y}$):
$w_{\pixel{X}\pixel{Y}}=\exp(-\beta{(x-y)}^2)$
measuring pixel intensity difference with parameter $\beta$ is applied. The probabilities $P(V_M)$ are set to 1 for the seeds of a particular label and 0 for the rest. This procedure is repeated for each label to obtain the corresponding probabilities $P(V_U)$. Finally, each unmarked node from $V_U$ receives the label with the highest probability among all labels.

\subsection{Weight Function}
\label{subsec:weight_func}
\begin{figure}[t]
\centering
    \includegraphics[width=0.49\textwidth]{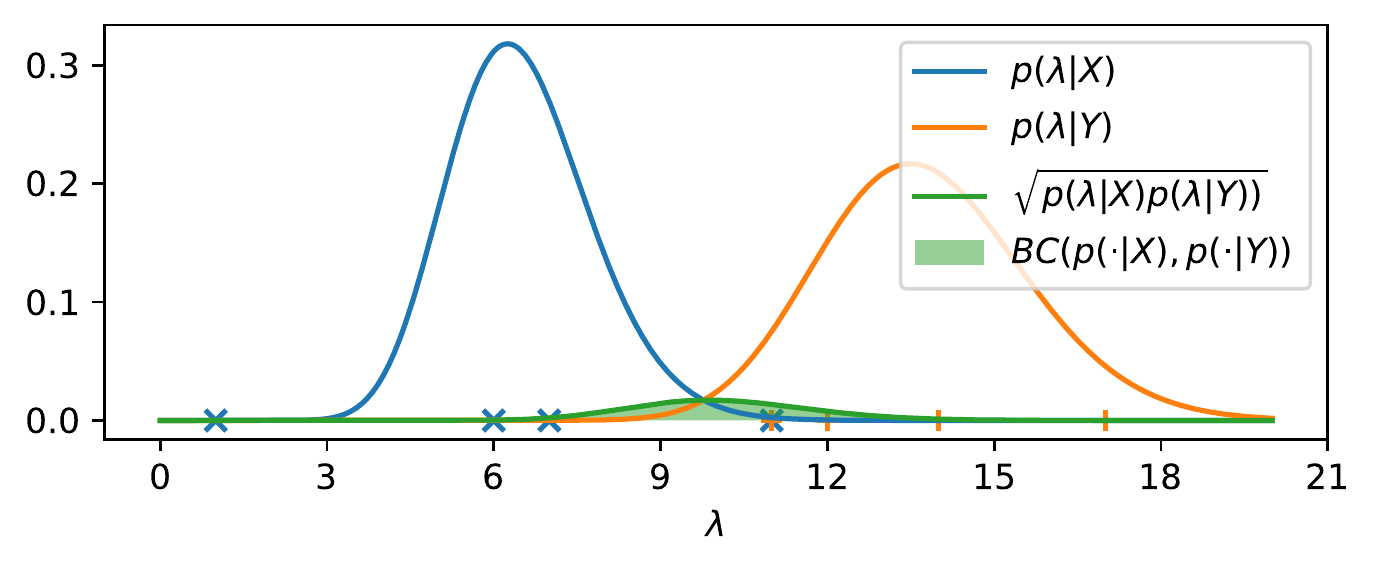}   \includegraphics[width=0.49\textwidth]{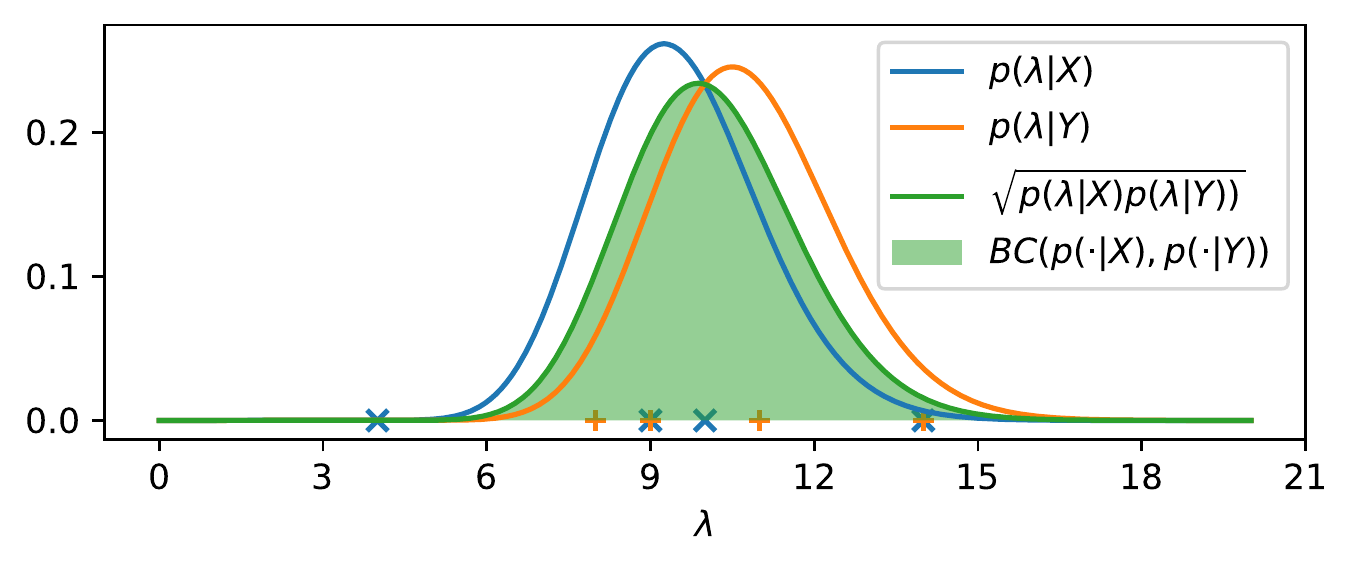}
    \caption{
    Illustration of the behavior of the Bhattacharyya coefficient as a similarity measure for weight function applied to two measurement sets under Poisson noise.
    }\label{fig:weight_function_illustration}
\end{figure}

In general, the edge weight function $w_{\pixel{X}\pixel{Y}}$ should express a similarity of the two adjacent pixels $\pixel{X}$ and $\pixel{Y}$.
For this we model the value of a pixel $\pixel{X}$ as a probability distribution $p(x|\parameters{X})$ with parameters $\parameters{X}$.
The actual pixel values in an image are thus assumed to be drawn from the (per-pixel) distribution.
Let $X$ and $Y$ be multi-sets of $n$ samples each, drawn from the distributions of the adjacent pixels $\pixel{X}$ and $\pixel{Y}$.
Since we only have one sample per pixel (i.e., the \textit{actual} image value) we assume that pixels in the \textit{neighborhood} are from the same distribution.
This assumption is also made in previous work~\cite{ang2016statistical,ang2016ttest}.
\autoref{sec:neighborhood} is concerned with the construction of these neighborhoods. We can estimate the distribution of the parameters $\parameters{X}$ given $X$ (and $\parameters{Y}$ given $Y$ accordingly) via Bayesian estimation:
\begin{equation}
   \label{eq:param_given_set}
    p(\parameters{X} | X) = \frac{p(X | \parameters{X}) p(\parameters{X})} {\int_{P_\kappa} p(X | \kappa) p(\kappa) d\kappa}
    = \frac{p(\parameters{X}) \prod_{x \in X} p(x | \parameters{X})} {\int_{P_\kappa} p(\kappa) \prod_{x \in X} p(x | \kappa) d\kappa}
    = \frac{\prod_{x \in X} p(x | \parameters{X})} {\int_{P_\kappa} \prod_{x \in X} p(x | \kappa) d\kappa}
\end{equation}
Here, we first applied Bayes' theorem and then used the fact that all samples in $X$ are independent.
In the last step we assumed that $\kappa$ is uniformly distributed in $P_\kappa$, i.e., we assume no further prior knowledge about its distribution within $P_\kappa$.
In this form, $p(\parameters{X} | X)$ is easy to apply to many noise models since it only depends on the PDF $p(x | \kappa)$.
Given $X$ and $Y$, we can then define $w_{\pixel{X}\pixel{Y}} := S(p(\cdot | X), p(\cdot | Y))$ using a similarity measure $S$ between probability distributions.
In this paper, we use the Bhattacharyya coefficient~\cite{bhattacharyya1946measure} $BC(p(\cdot),q(\cdot)) = \int_P \sqrt{p(r)q(r)} dr$, which is a closed form similarity measure of probability distributions based on their PDF.
As illustrated in \autoref{fig:weight_function_illustration} it is 0 for non-overlapping PDFs and increases with the amount of overlap up to a value of 1.
It thus enables graphical interpretation of the weight function and it is easily applicable to any noise model that has a PDF.
Using \autoref{eq:param_given_set} it allows for simplifications:

\begin{equation}
\begin{aligned}[b]
   \label{eq:framework}
    w_{\pixel{X}\pixel{Y}} &= BC(p(\cdot | X), p(\cdot | Y)) = \int_{P_\kappa} \sqrt{p(\kappa | X) p(\kappa | Y)} d\kappa\\
                   &= \int_{P_\kappa} \sqrt{\frac{\prod_{x \in X} p(x | \kappa)\prod_{y \in Y} p(y | \kappa)} {\int_{P_\kappa} \prod_{x \in X} p(x | \tilde{\kappa}) d\tilde{\kappa} \int_{P_\kappa} \prod_{y \in Y} p(y | \tilde{\kappa}) d\tilde{\kappa}}} d\kappa\\
                   &= \frac{\int_{P_\kappa} \sqrt{\prod_{x \in X} p(x | \kappa) \prod_{y \in Y}p(y | \kappa)}d\kappa}{\sqrt{\int_{P_\kappa} \prod_{x \in X} p(x | \kappa) d\kappa \int_{P_\kappa} \prod_{y \in Y} p(y | \kappa) d\kappa}}
\end{aligned}
\end{equation}
The final result is only dependent on the pixel-value PDF of the noise model.
The integrals in the numerator and denominator are quite similar and can be solved in similar fashion in practice (see \autoref{sec:noise_models}).

\subsection{Neighborhood}
\label{sec:neighborhood}
\begin{figure}[t]
    \begin{minipage}{.45\textwidth}
    \centering
    \includegraphics[width=0.585\columnwidth]{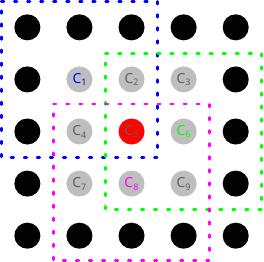}
    \caption{For every pixel, select neighborhood of highest probability with \autoref{eq:neighbor}. Only 3 out of $9=(2k+1)^2$, with $k=1$, (with centers $C_1, \ldots, C_9$) possible neighborhoods are drawn here.}\label{fig:neighbor}
    \end{minipage}
    \hspace{.02\textwidth}
    \begin{minipage}{.5\textwidth}
    \centering
    \includegraphics[width=0.6\columnwidth]{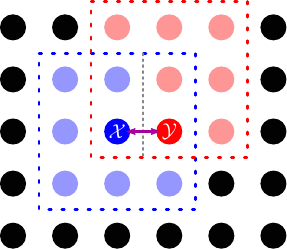}
    \caption{Solve overlap of neighborhoods by sorting and dividing by difference of Euclidean distance to both pixels. Pixels $\pixel{X}$, $\pixel{Y}$ are to be compared, blue/red dashed lines show respective selected neighborhoods.
    }\label{fig:overlap}
    \end{minipage}
\end{figure}

In this section we describe how to determine a fitting \textit{neighborhood} as needed in \autoref{subsec:weight_func}. In \cite{ang2016ttest} Bian et al.\ proposes a variation of a well proven (e.g., \cite{yokoya1989range}) optimal neighborhood selection schema which we also adopt here:
To determine the optimal neighborhood of $\pixel{X}$, let $X^1, \ldots, X^{(2k+1)^2}$ be the $(2k+1)^2$ quadratic neighborhoods including $\pixel{X}$ as shown in \autoref{fig:neighbor}.
The optimal neighborhood can now be found as the neighborhood that maximizes the probability of having sampled the pixel value $x$ of $\pixel{X}$. Formally, this is described as:
\begin{equation}\label{eq:neighbor}
	X = \underset{N\in\{X^1, \dots, X^{(2k+1)^2}\}}{\text{argmax}} p(x | N),
\end{equation}
where $p( \cdot | N)$ is the PDF of the assumed noise model with parameters estimated from the pixel intensities in $N$.
It should be noted that the neighborhoods $X$ and $Y$ of pixels $\pixel{X}$ and $\pixel{Y}$ can overlap.
In order to ensure statistical independence of samples in $X$ and $Y$, pixels in the overlap have to be assigned to either $X$ or $Y$.
In \cite{ang2016ttest} it is proposed to assign pixels from the overlap based on the Euclidean distance to one of the center pixels.
However, this ultimately leads to a non-symmetric weight function.
Instead, we sort and divide using the difference of the Euclidean distances to \textit{both} center pixels, which results in a dividing line orthogonal to the graph edge (see \autoref{fig:overlap}).
Ties are resolved deterministically.

\section{Application to Concrete Noise Models}
\label{sec:noise_models}

In this section we apply the framework presented in \autoref{sec:framework} to three noise models relevant to various imaging modalities and obtain closed form solutions. We choose Poisson noise as it is very common in biology \cite{risse2013fim,micronoise2006Vonesch}, as well as Gaussian noise - with two different configurations: fixed variance over the whole image and variable variance per image region - which is common in many images including medically relevant techniques like MRI and CT. For Poisson noise this is the first derivation of a noise model specific weight function for random walker, whereas the two versions of Gaussian noise model have been considered in \cite{ang2016statistical,ang2016ttest}, respectively. See Appendix \autoref{supsec:noise_models}
for derivations in more detail.

\subsection{Poisson Noise}
In this subsection we assume an image that is affected by Poisson noise (also called ``shot noise``), which occurs, for example, as an effect of photon or electron counting in imaging systems (e.g., fluorescence microscopy \cite{micronoise2006Vonesch}, positron emission tomography \cite{petnoise1982Shepp}, low dose CT \cite{ctnoise2001Lu}).
In this model the measured pixel values $x,y \in \mathbb{N}$ are drawn from a Poisson distribution with unknown parameter $\kappa = \lambda$, which corresponds to the true pixel value.
We assume the prior distribution of $\lambda$ to be uniform in $P=(0, a)$ for some sufficiently large value $a$.
For convenience, we define $\sums{X}=\sum_{x \in X} x$ and $\sums{Y}=\sum_{y \in Y} y$.
Then, using \autoref{eq:framework}, we obtain:
\begin{equation}
\begin{aligned}[b]
    w_{\mathcal{X}\mathcal{Y}}=BC(p(\cdot | X), p(\cdot | Y))
    = &\frac{\sqrt{\int_P \prod_{x\in X} \frac{e^{-\lambda}\lambda^{x}}{x!}\prod_{y\in Y}\frac{e^{-\lambda}\lambda^{y}}{y!}}d\lambda}{\sqrt{\int_P \prod_{x\in X} \frac{e^{-\lambda}\lambda^{x}}{x!} d\lambda \int_P \prod_{y\in Y} \frac{e^{-\lambda}\lambda^{y}}{y!} d\lambda}}\\
    = &\frac{\Gamma(\frac{\sums{X}+\sums{Y}}{2} + 1)}{\sqrt{\Gamma(\sums{X}+1)\Gamma(\sums{Y}+1)}} \label{eq:poisson_gamma}
\end{aligned}
\end{equation}
Strictly speaking, \autoref{eq:poisson_gamma} only holds asymptotically for $a \to \infty$. Since we assume no prior knowledge about the distribution of $\lambda$, we can let $a$ tend to infinity to use \autoref{eq:poisson_gamma} as the weight function. The convergence of \autoref{eq:poisson_gamma} is discussed in more detail in the Appendix \autoref{sec:A_poisson}.

\subsection{Multivariate Gaussian Noise with Constant Covariance}\label{sec:gauss}

In this subsection we assume an m-channel image with concrete pixel values $x,y \in \mathbb{R}^m$.
The true image values are perturbed by additive Gaussian noise and are thus modeled by a Gaussian PDF with (unknown) parameter $\kappa = \mu \in \mathbb{R}^m$, which also corresponds to the true pixel value.
This noise model applies, for example, in complex valued MRI images \cite{mrinoise1985henkelman}.
Further, we assume that $\mu$ is priorly uniformly distributed in $P=(-a, a)^m$ for some sufficiently large value $a$.
The covariance matrix $C$ is assumed to be constant for the whole image.
It should be noted that the special case $m=1$ is the setting that is assumed in prior work~\cite{ang2016statistical}.
Starting from \autoref{eq:param_given_set} we obtain:
\begin{equation}
\begin{aligned}[b]
   \label{eq:mdim_gauss_pdf1}
    p(\mu | X) = &\frac{\prod_{x \in X} \frac{1}{\sqrt{(2\pi)^m \det(C)}} \exp(-\frac{1}{2} (x - \mu)^T C^{-1} (x - \mu))} {\int_P \prod_{x \in X} \frac{1}{\sqrt{(2\pi)^m \det(C)}} \exp(-\frac{1}{2} (x - \tilde{\mu})^T C^{-1} (x - \tilde{\mu})) d\tilde{\mu}}\\
                = &\frac{\exp(-\frac{1}{2} \sum_{x \in X} (x - \mu)^T C^{-1} (x - \mu))} {\int_P \exp(-\frac{1}{2} \sum_{x \in X} (x - \tilde{\mu})^T C^{-1} (x - \tilde{\mu})) d\tilde{\mu}}\\
                = &\frac{1}{\sqrt{(2\pi)^m \det(\frac{C}{n})}} \exp\bigg(-\frac{1}{2} \big(\mu - \sum_{x\in X} \frac{x}{n}\big)^T \left(\frac{C}{n}\right)^{-1} \big(\mu - \sum_{x \in X} \frac{x}{n}\big)\bigg)
\end{aligned}
\end{equation}
Similar to \autoref{eq:poisson_gamma}, this equation only holds asymptotically for $a\to \infty$, but we can choose $a$ large enough for arbitrary precision.
\autoref{eq:mdim_gauss_pdf1} shows that $p(\mu | X)$ is simply the density function of a normal distribution with mean $\sum_{x\in X} \frac{x}{n} =: \bar{X}$ and covariance matrix $\frac{C}{n}$ (and $p(\mu|Y)$ accordingly).
As shown in \cite{nielsen2011burbea} the Bhattacharyya coefficient is then:
\begin{equation}\label{eq:mult_gauss_fin}
    w_{\mathcal{X}\mathcal{Y}} = BC(p(\cdot | X), p(\cdot | Y)) \propto \exp\bigg(-\frac{1}{8}(\bar{X}- \bar{Y})^T\left(\frac{C}{n}\right)^{-1}(\bar{X} - \bar{Y})\bigg)
\end{equation}
\subsection{Gaussian Noise with Signal-Dependent Variance}

In this subsection we assume additive Gaussian noise on single channel images where, however, $\sigma^2$ differs between image regions (in contrast to \autoref{sec:gauss} which assumes a global, constant $C=\sigma^2$ for m=1).
Thus, pixel values are modeled by $x = \mu_\pixel{X} + \mathcal{N}(0, \sigma_\pixel{X}^2)$.
A special case is Loupas noise, where $\sigma_\pixel{X}^2 = \mu_\pixel{X}\sigma^2$ for some fixed (global) $\sigma^2$. It applies, for example, to speckled SAR and medical ultrasound images~\cite{ang2016ttest,TENBRINCK2015arbnoise}
Thus, we have to estimate $\mu$ and $\sigma^2$ simultaneously and set $\kappa$ from \autoref{eq:framework} to be $(\mu, \sigma^2)$.

To solve \autoref{eq:framework} we set $P_a := (0, a)\times(-a,a)$ (again, $a$ sufficiently large) and assume the prior distribution of $(\mu, \sigma^2)$ to be uniform.
We then have to calculate the integrals in the enumerator and denominator of \autoref{eq:framework}, which can be reformulated to a similar form and then can be calculated analogously:
\begin{equation}
\begin{aligned}[b]
    & \int_P \left(\frac{1}{\sqrt{2\pi \sigma^2}}\right)^n\exp\left(-\frac{1}{2\sigma^2}\sum_{x\in X} (x - \mu)^2)\right) d(\mu,\sigma^2) \\ \label{eq:easy_integral0}
    = & \frac{1}{\sqrt{n}} \left(\frac{1}{2\pi}\right)^{n-1} \Gamma\left(\frac{n-3}{2}\right)\left(\frac{1}{4n}\sum_{x_1, x_2 \in X} (x_1 - x_2)^2 \right)^{\frac{-n+3}{2}} \\
\end{aligned}
\end{equation}%
\begin{equation}
\begin{aligned}[b]
    & \int_P \left(\frac{1}{\sqrt{2\pi \sigma^2}}\right)^n\exp\left(-\frac{1}{4\sigma^2}\sum_{z\in X\cup Y} (z - \mu)^2)\right) d(\mu,\sigma^2) \\
    = & \frac{1}{\sqrt{n}} \left(\frac{1}{2\pi}\right)^{n-1} \Gamma\left(\frac{n-3}{2}\right)\left(\frac{1}{16n}\sum_{z_1, z_2 \in X\cup Y} (z_1 - z_2)^2 \right)^{\frac{-n+3}{2}}\label{eq:easy_integral2}
\end{aligned}
\end{equation}
Inserting these equations into \autoref{eq:framework} and canceling the fraction yields:
\begin{equation}
\begin{aligned}[b]
    \label{eq:bc_variable_gaussian_prelim}
    w_{\mathcal{X}\mathcal{Y}} = BC(p(\cdot|X), p(\cdot|Y)) &= \left( 4 \frac{\sqrt{\sum_{x_1, x_2 \in X}(x_1 - x_2)^2 \sum_{y_1, y_2 \in Y}(y_1 - y_2)^2}}{\sum_{z_1, z_2\in X\cup Y}(z_1 - z_2)^2} \right)^{\frac{n-3}{2}} \\
    &= \left(\frac{\sqrt{Var(X)Var(Y)}}{Var(X\cup Y)} \right)^{\frac{n-3}{2}}
\end{aligned}
\end{equation}

\section{Experimental Results}\label{sec:results}
We conduct three experiments to compare our suitable methods with Grady~\cite{grady2006random} and the noise model based approaches of Bian et al.~\cite{ang2016statistical,ang2016ttest}:
We demonstrate differences in seed propagation on synthetic data under different noise conditions and report results on real world image data (FIM~\cite{risse2013fim} and MRI~\cite{knoll2020fastmri}).
To compare to Grady~\cite{grady2006random}, we use two approaches:
1) We search for the on average best $\beta$ over the whole dataset.
This mimics the behavior of a user determining a ``good'' value for $\beta$ initially on a some images.
2) We also report the results where $\beta$ is tuned optimally for every single image and label configuration, which is unrealistic, but serves as a performance upper bound.
\begin{figure}[t]
    \hspace{7mm}
    \begin{minipage}{.15\textwidth}
    \begin{center}
    \begin{tabular}{c}
      \includegraphics[width=1\linewidth,valign=m]{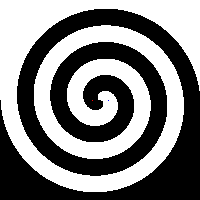}\\
      (a)\\
      \includegraphics[width=1\linewidth,valign=m]{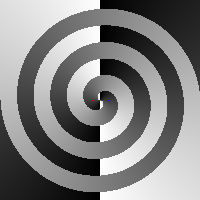}\\
      (b)\\
    \end{tabular}
    \end{center}

    \end{minipage}
    \begin{minipage}{.05\textwidth}
    \hspace{3mm}
    \end{minipage}
    \begin{minipage}{.6\textwidth}
    \centering
    \begin{tabular}{cccc}
      & Poisson & Loupas & 2D Gaussian \\
      \rotatebox[origin=c]{90}{Grady~\cite{grady2006random}}& \includegraphics[width=.2\linewidth,valign=m]{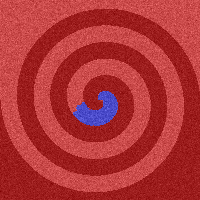} & \includegraphics[width=.2\linewidth,valign=m]{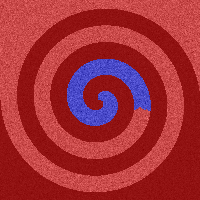} & \includegraphics[width=.2\linewidth,valign=m]{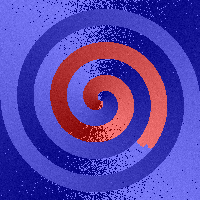}\\
    \rotatebox[origin=c]{90}{Bian~\cite{ang2016ttest}} & \includegraphics[width=.2\linewidth,valign=m]{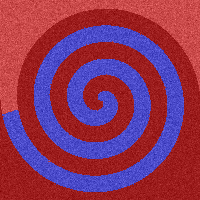} & \includegraphics[width=.2\linewidth,valign=m]{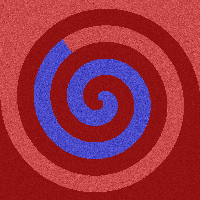} & (Not applicable)\\
    \rotatebox[origin=c]{90}{Ours} & \includegraphics[width=.2\linewidth,valign=m]{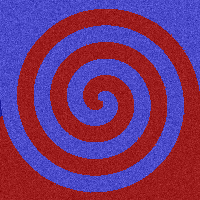} & \includegraphics[width=.2\linewidth,valign=m]{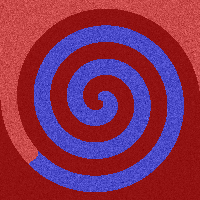} & \includegraphics[width=.2\linewidth,valign=m]{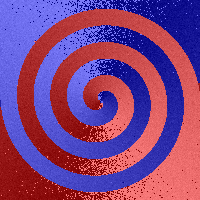}\\
    \end{tabular}
    \end{minipage}
    \caption{
       Illustration of behavior on spiral synthetic data with mediocre noise level ($\lambda_0/\lambda_1 = 32/64$, $\sigma=0.2$ (Loupas), $\sigma=0.3$ (Gaussian)).
       (a) Ground truth image for all cases;
       (b) Phase of the uncorrupted 2D vector-valued image, discontinuities are due to phase wraps, the magnitude was set to be constant at 1.
    }\label{fig:illustration_spiral}
\end{figure}
\begin{figure}[t]
    \includegraphics[width=0.32\columnwidth]{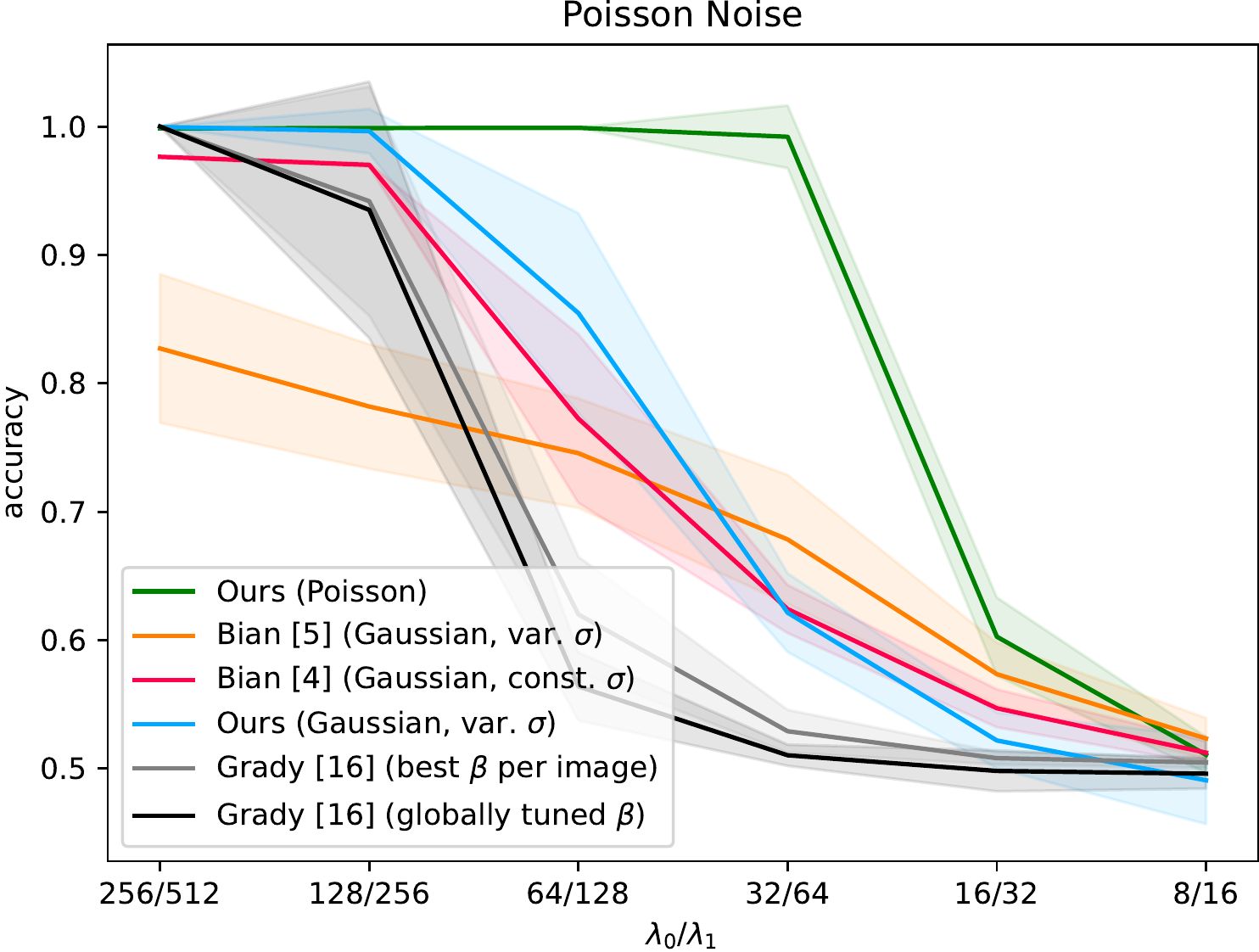}
    \includegraphics[width=0.32\columnwidth]{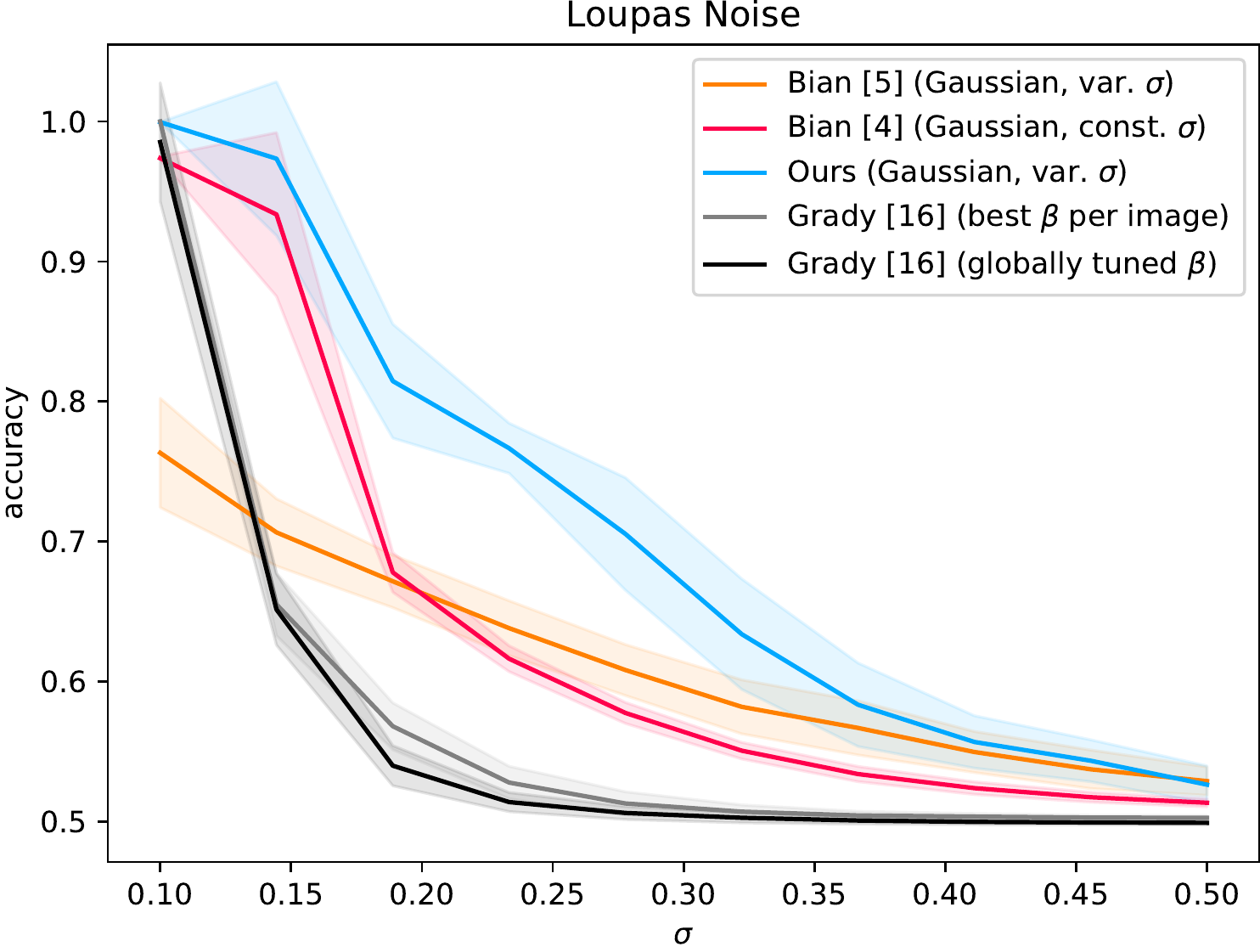}
    \includegraphics[width=0.32\columnwidth]{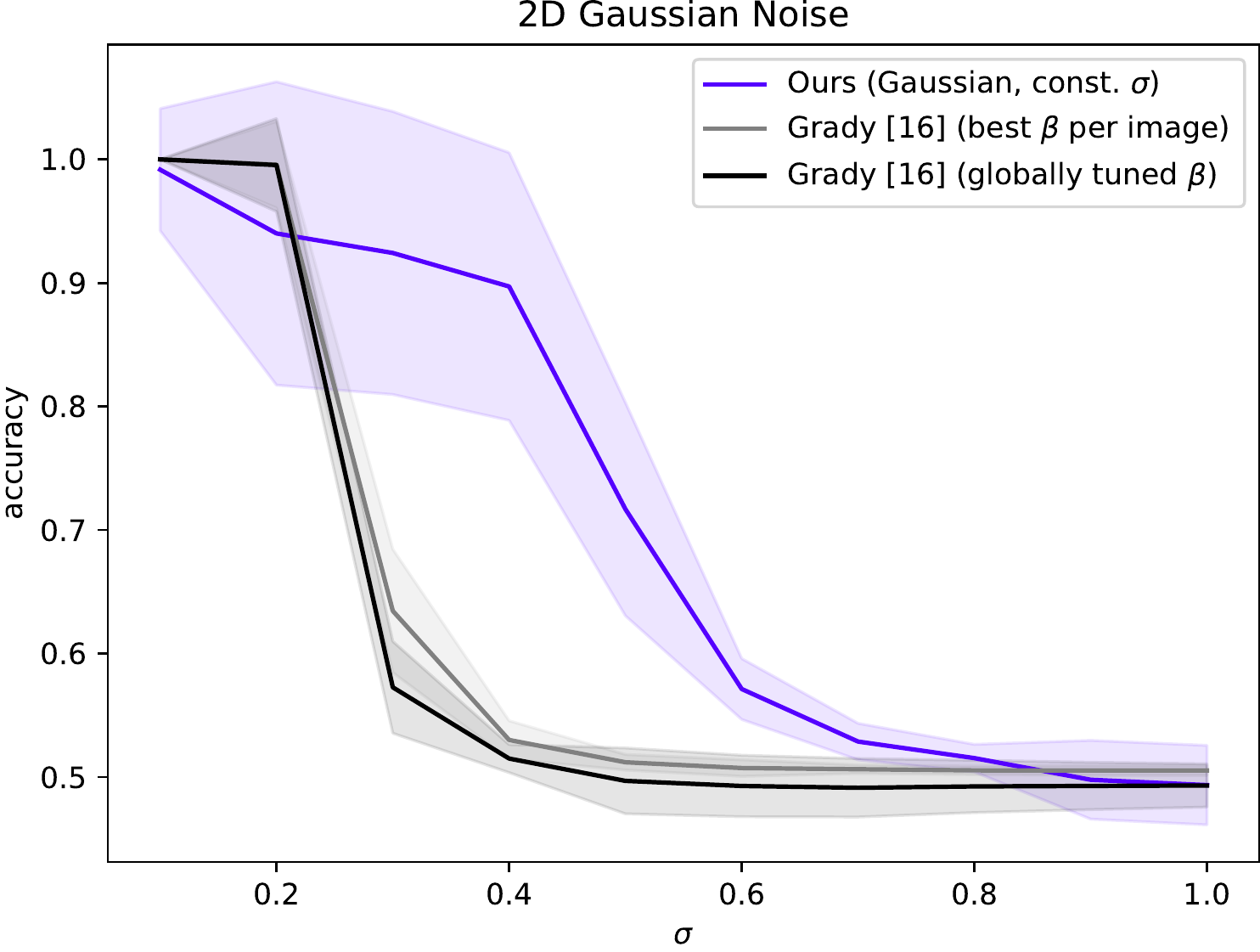}
    \caption{
       Accuracy of applicable methods on synthetic image for Poisson/Loupas/2D Gaussian noise (left/center/right). Lines show mean, shadows show standard deviation.
    }\label{fig:accuracy_synthetic}
\end{figure}

\subsection{Results on Synthetic Data}
To evaluate how class probabilities propagate from an initial seed under noisy conditions, we follow Grady's idea~\cite{grady2006random} and generate a spiral structure, see \autoref{fig:illustration_spiral}.
For scalar images, the two regions simply differ in intensity. 
For the (2D) vector-valued image the two regions correspond to vector fields defining a flow into and out of the spiral, respectively.
The magnitude of all vectors is unity.
For both regions, one seed is placed at the central start of the spiral.
In a total of three noise scenarios, the base images are perturbed by Poisson and Loupas noise (scalar) as well as uncorrelated, symmetric 2D-multivariate Gaussian noise (vector image) which can also be interpreted as complex Gaussian noise as it would be present in MRI images.
For each of the three scenarios, applicable weight functions are applied and 100 realizations of the random noise for each noise type and each noise level were evaluated.
For Poisson noise, we modulate the noise level by decreasing the mean intensity of the two image regions ($\lambda_0$, $\lambda_1$) from (256, 512) to (8, 16), which reduces the signal-to-noise ratio.
For Loupas noise, the region intensity were set to 0.1 and 1 and $\sigma$ varied between 0.1 and 0.5.
For the vector images the uncorrelated, symmetric 2D-multivariate Gaussian noise $\sigma$ was set in $[0, 1]$.
\autoref{fig:illustration_spiral} examplarily shows the results on images with intermediate noise levels, while \autoref{fig:accuracy_synthetic} shows quantitative results.
Accuracy is a suitable measure in this scenario since the two classes are of equal size.

All three scenarios show that our method with the suitable noise model leads to superior performance, which gets beat substantially only in one scenario (\cite{grady2006random} for low level of multivariate Gaussian noise).
This however still has the drawback of having to choose the correct parameter $\beta$ and it drops off at a noise level of roughly $\sigma = 0.3$.
For Poisson noise, the hardest competitors are \cite{ang2016ttest} and our method for variable Gaussian noise, which is unsurprising, since Poisson noise can be approximated by Gaussian noise with signal-dependent $\sigma$.
\cite{ang2016ttest} also works well for strong Loupas noise, which is also the noise model it was designed for, however still falls short to the proposed method at all noise levels.

\subsection{Results on Real World Data}
\label{subsec:eval_meth}
\begin{figure}[t]
    \begin{center}
    \begin{tabular}{cccccc}
        \includegraphics[width=.16\linewidth,valign=m]{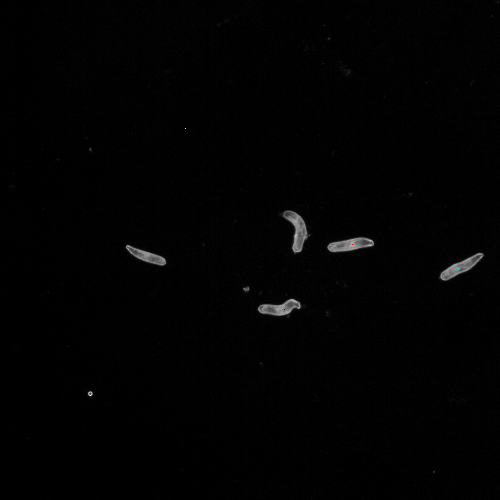} &
        \includegraphics[width=.16\linewidth,valign=m]{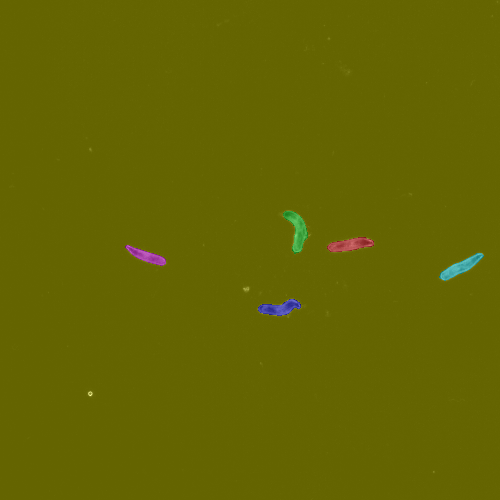} &
        \includegraphics[width=.16\linewidth,valign=m]{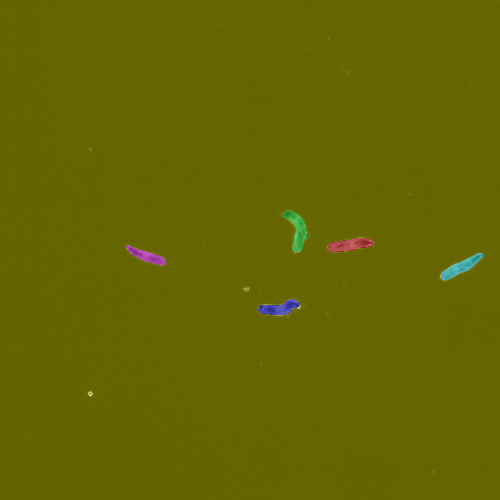} &
        \includegraphics[width=.16\linewidth,valign=m]{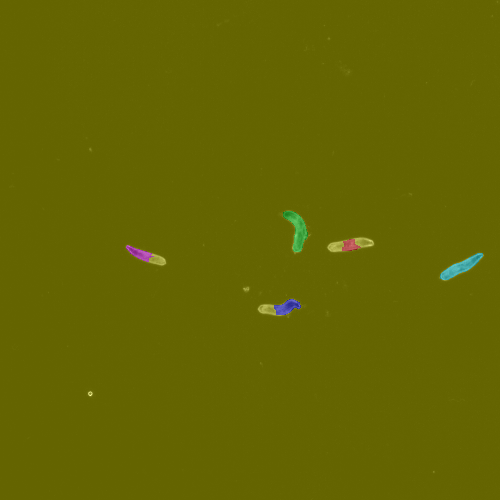} &
        \includegraphics[width=.16\linewidth,valign=m]{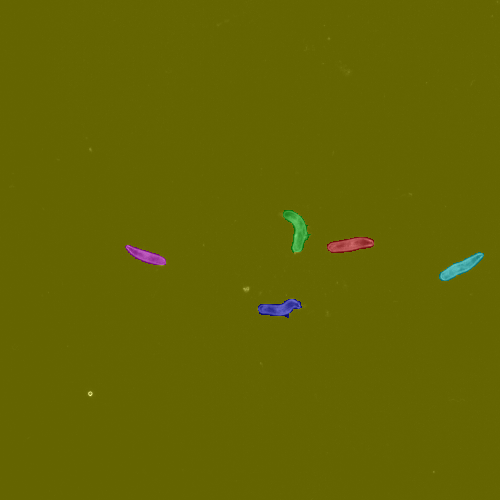} &
        \includegraphics[width=.16\linewidth,valign=m]{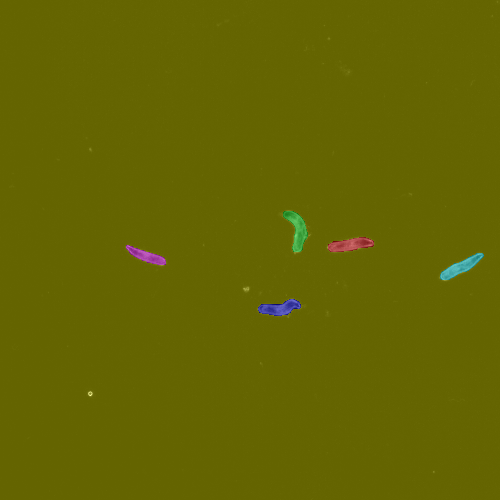}\\
        Image \& seeds &
        GT &
        Grady~\cite{grady2006random} &
        Bian~\cite{ang2016ttest} &
        Ours, var. $\sigma$ &
        Ours, Poisson\\
    \end{tabular}
    \end{center}
    \begin{center}
    \begin{tabular}{l|c|c|c|c}
	    	& Grady~\cite{grady2006random} & Bian~\cite{ang2016ttest} & Ours (var. $\sigma$) & Ours (Poisson) \\
	    \hline
	    VOI & 0.013 & 0.039 & 0.025 & 0.010 \\
	    \hline
	    ARAND  & 0.038 & 0.185 & 0.075 & 0.026 \\ \hline
    \end{tabular}
    \end{center}
    \caption{
       Qualitative results of FIM Larvae dataset experiments with initial seeds and class maps overlaid on top of the original image.
       Quantitative results for the shown images are reported in table.
       For Grady the best $\beta$ for this image was selected.
    }\label{fig:illustration_fim}
\end{figure}

\begin{figure}[t]
\begin{center}
    \begin{tabular}{cccccc}
        \includegraphics[width=.16\linewidth,valign=m]{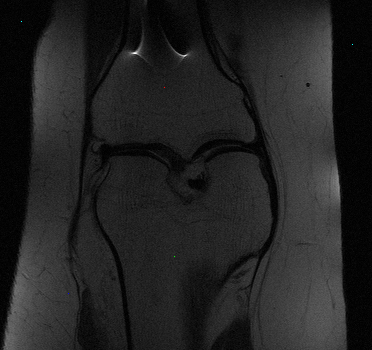} & \includegraphics[width=.16\linewidth,valign=m]{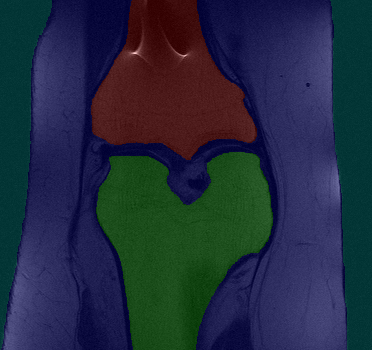} &
        \includegraphics[width=.16\linewidth,valign=m]{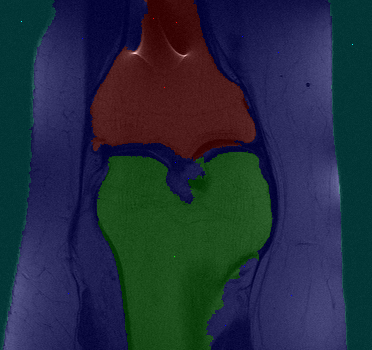} &
        \includegraphics[width=.16\linewidth,valign=m]{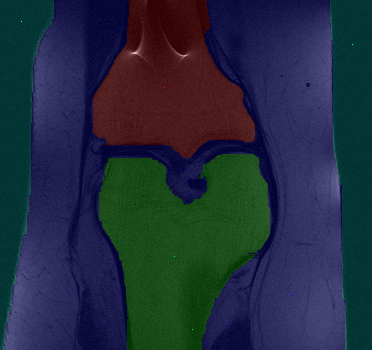} &
        \includegraphics[width=.16\linewidth,valign=m]{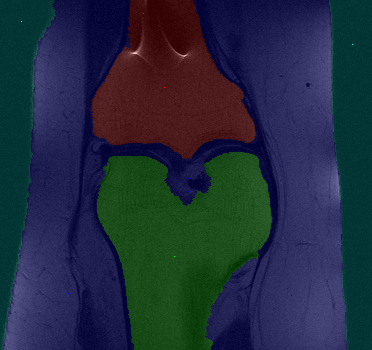} &
        \includegraphics[width=.16\linewidth,valign=m]{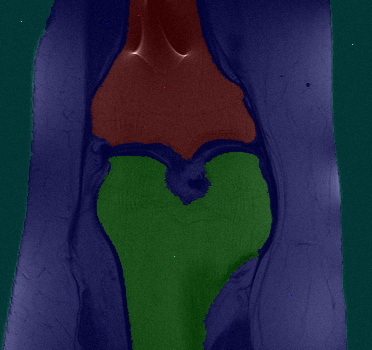}\\
        Image \& seeds &
        GT &
        Grady~\cite{grady2006random}&
        Bian~\cite{ang2016statistical}&
        Ours, const. $\sigma$&
        Ours, const. $\sigma$\\
        &
        &
        on $||\cdot||_2$ &
        on $||\cdot||_2$ &
        on $||\cdot||_2$ &
        on 2D\\
    \end{tabular}
    \end{center}
    \begin{center}
    \begin{tabular}{l|c|c|c|c}
	    	& Grady~\cite{grady2006random}, $||\cdot||_2$ & Bian~\cite{ang2016statistical}, $||\cdot||_2$ & Ours (const. $\sigma$, $||\cdot||_2$) & Ours (const. $\sigma$, 2D) \\
	    \hline
	    VOI & 0.48 & 0.37 &  0.30 & 0.26 \\
	    \hline
	    ARAND & 0.11 & 0.079 & 0.063 & 0.051 \\ \hline
    \end{tabular}

    \end{center}
    \caption{
       Qualitative results of fastMRI dataset experiments with ten additional seeds and class maps overlaid on top of the original image.
       Quantitative results for the shown images are reported in table.
       For Grady the best $\beta$ for this image was selected.
    }\label{fig:illustration_mri}
\end{figure}

For the following real world data evaluation we employ an automatic incremental seed placement strategy~\cite{mcguinness2011automated}:
Initially, all connected components of individual classes are assigned a first seed point at their center.
Then, based on an intermediate result of the method to be evaluated, additional seeds are placed for the class with the worst Dice score in the largest misclassified region.
This allows observation of 1) mean segmentation quality for a given number of seeds and 2) the required number of seeds to achieve a specific quality.
Quality measurement for the following multi-class problems is done using the Variation of Information (VOI)~\cite{meila2003voi} and Adapted Rand Error (ARAND).
Both produce good results even in the presence of class imbalance.
VOI measures the mutual information of the two multi-class segmentations with 0 describing full and 1 no agreement.
ARAND is defined as one minus the Adjusted Rand Index(ARI)~\cite{rand1971objective}, which measures if pixel pairs are labeled accordingly in the segmentations. It is normalized to be 0 for random and 1 for perfectly matching segmentations, meaning ARI $<0$ ($\Rightarrow$ ARAND $>1$) is possible for worse-than-random segmentations.
For intuition on the scores see the examples in \autoref{fig:illustration_fim} and \autoref{fig:illustration_mri}.

\begin{figure}[t]
    \includegraphics[width=0.49\columnwidth]{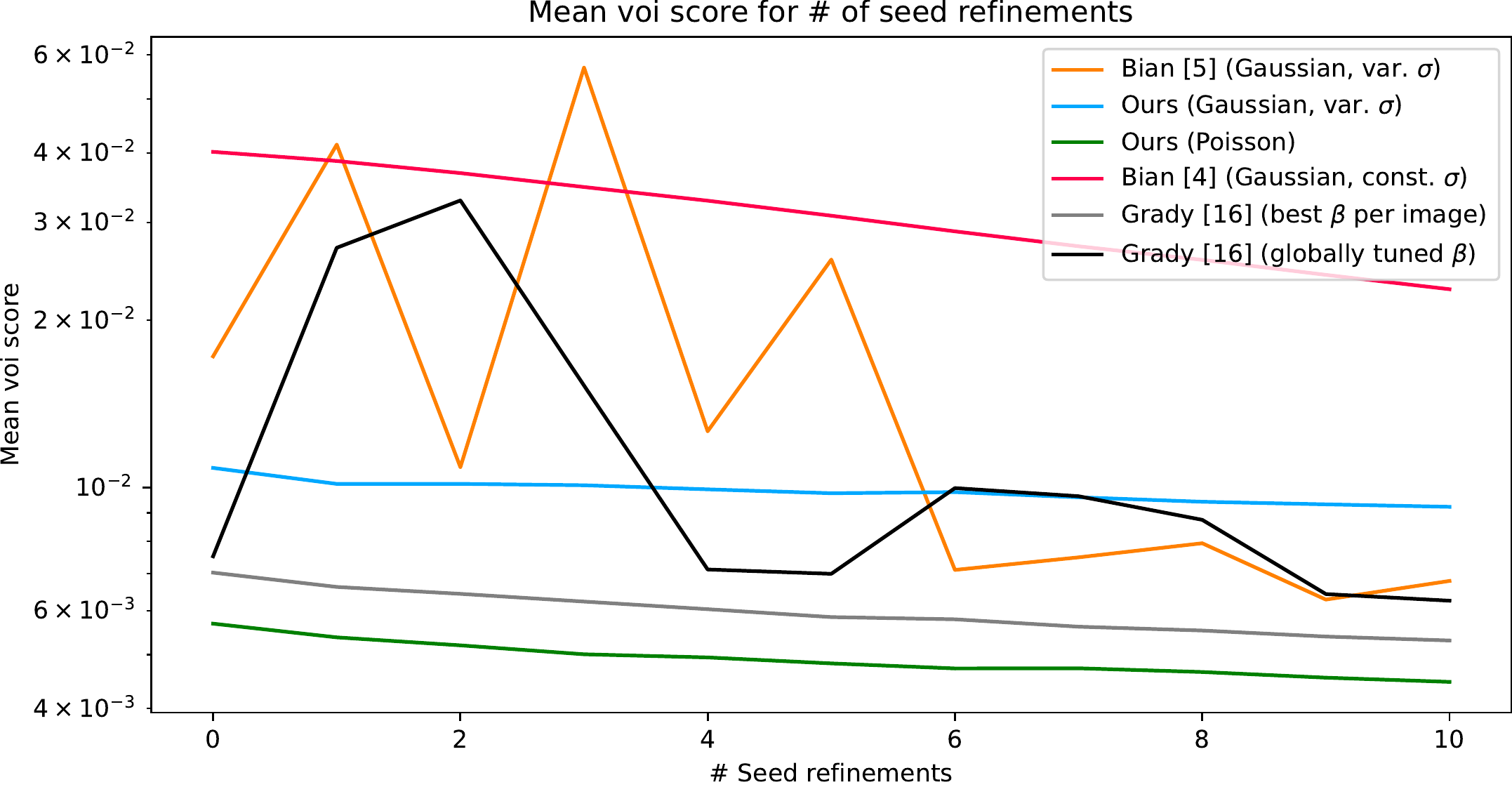}
    \includegraphics[width=0.49\columnwidth]{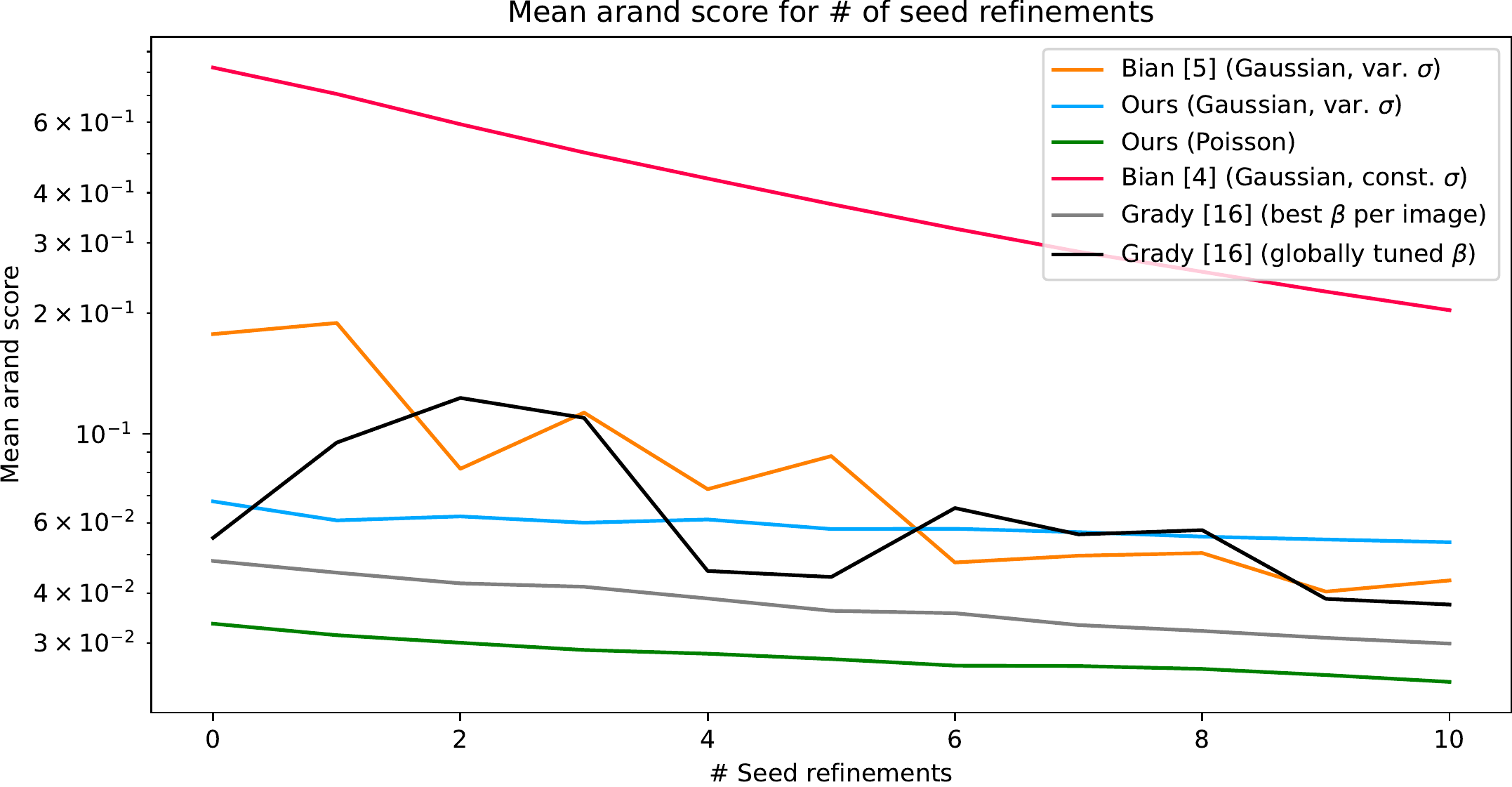}
    \includegraphics[width=0.49\columnwidth]{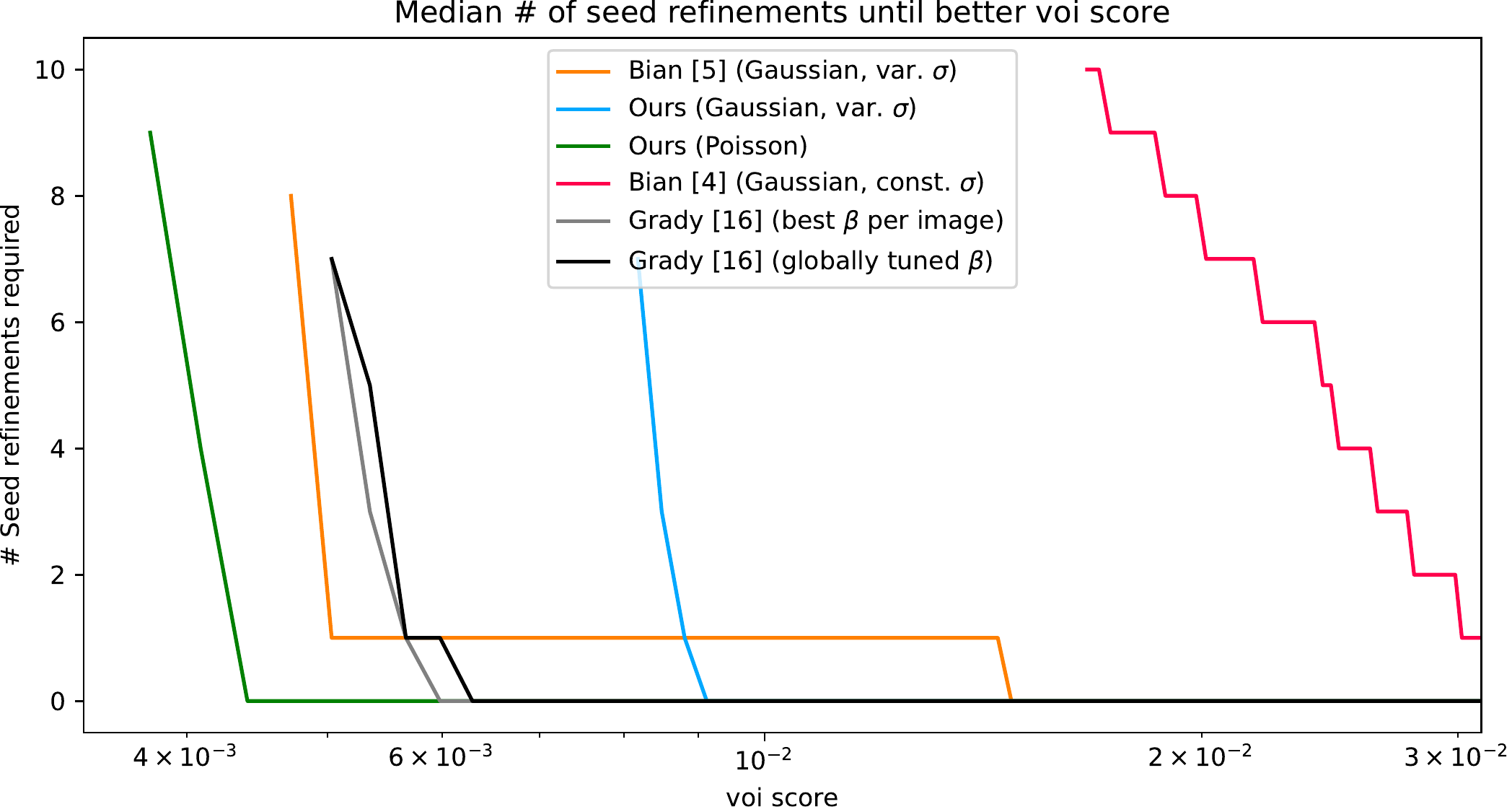}
    \includegraphics[width=0.49\columnwidth]{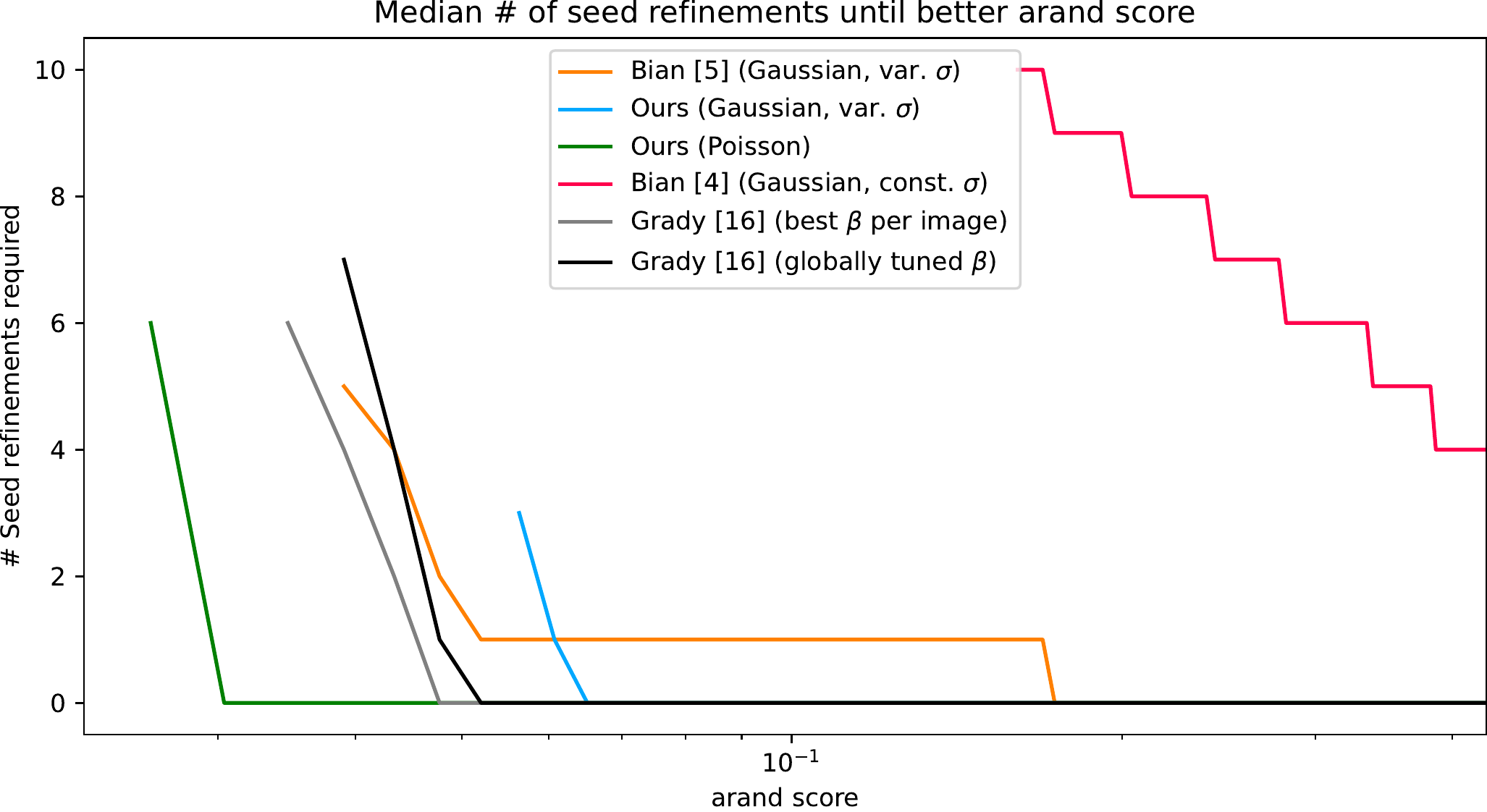}
    \caption{FIM Larvae dataset experiments:
       Mean VOI/ARAND scores after placing $n$ additional seeds (top) and median number of additional seeds required to achieve a specific VOI/ARAND score (bottom). Seeds were set as explained in \autoref{subsec:eval_meth}.
       }\label{fig:eval_quantitative_fim}
\end{figure}
\vspace{1mm}
\noindent
{\bf FIM larvae images.}
The FIM larvae dataset consists of 21 images of Drosophila Melanogaster larvae acquired by the FIM technique \cite{risse2013fim} where Poisson noise can be assumed.
The hand labeled ground truth masks consist of the background and up to five foreground classes: one per larva.
The images and corresponding ground truth masks are available online \footnote{\url{https://uni-muenster.sciebo.de/s/DK9F0f6p5ppsWXC}}.

The results in \autoref{fig:eval_quantitative_fim} show that the proposed weight function under the Poisson model outperforms all competitor methods.
Notably, it shows better scores without additional seeds than other methods achieve even with up to 10 additional seeds (see bottom row).
Other weight functions tend to either under-segment (Grady~\cite{grady2006random}, Bian et al.~\cite{ang2016ttest}) or over-segment (ours with variable variance Gaussian model) the larvae (see example in \autoref{fig:illustration_fim}).

\begin{figure}[t]
    \includegraphics[width=0.49\columnwidth]{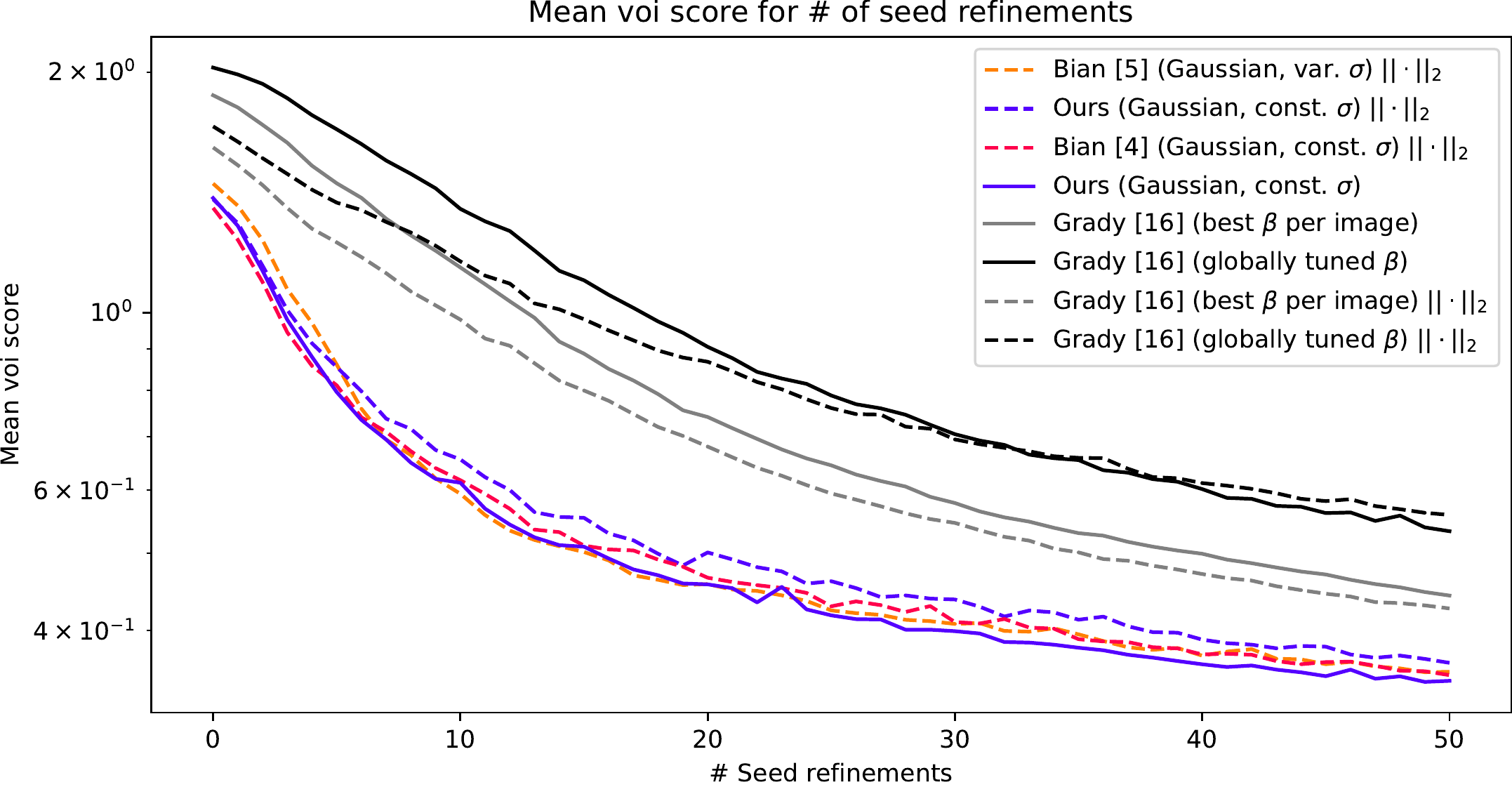}
    \includegraphics[width=0.49\columnwidth]{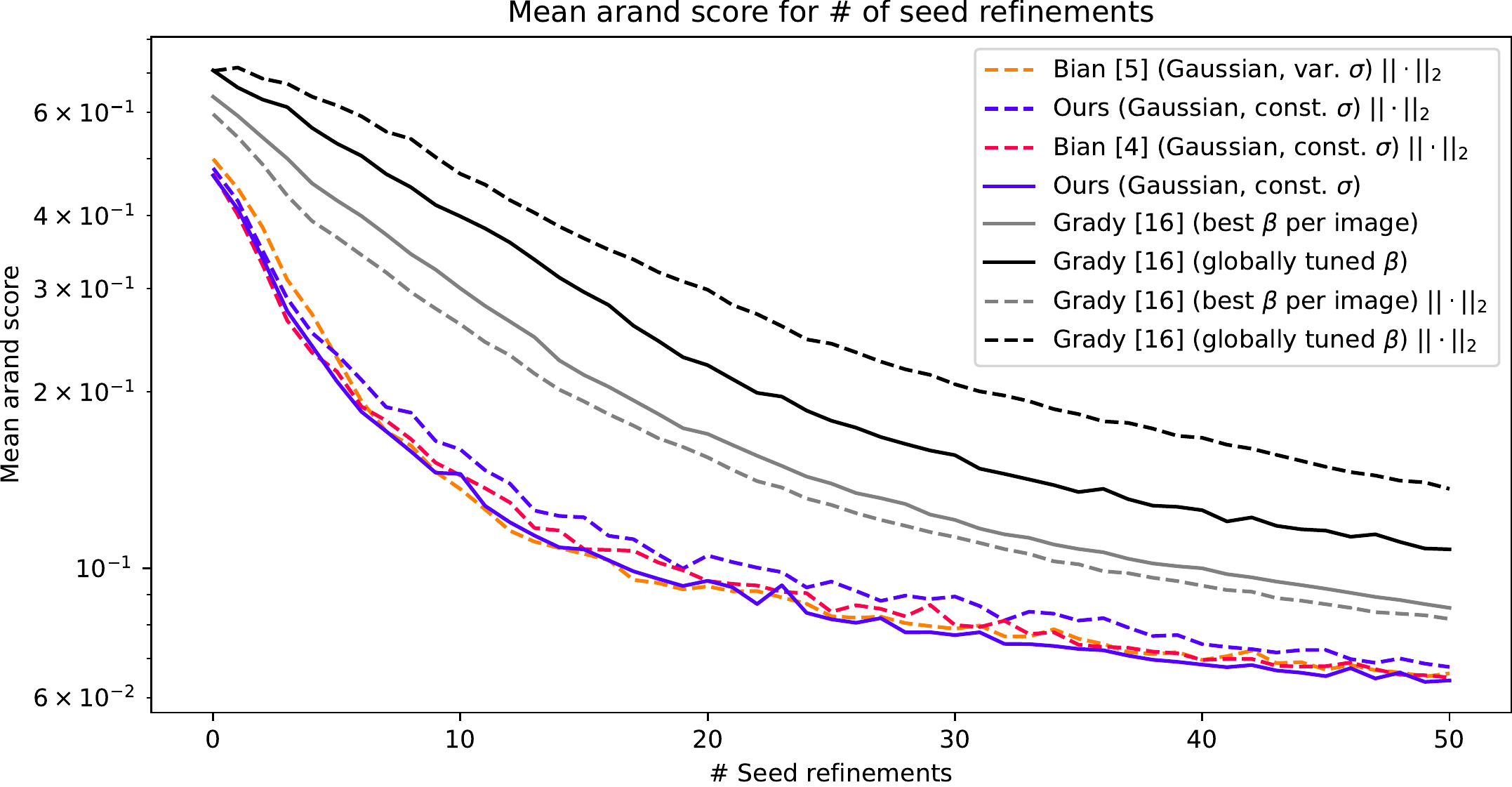}
    \includegraphics[width=0.49\columnwidth]{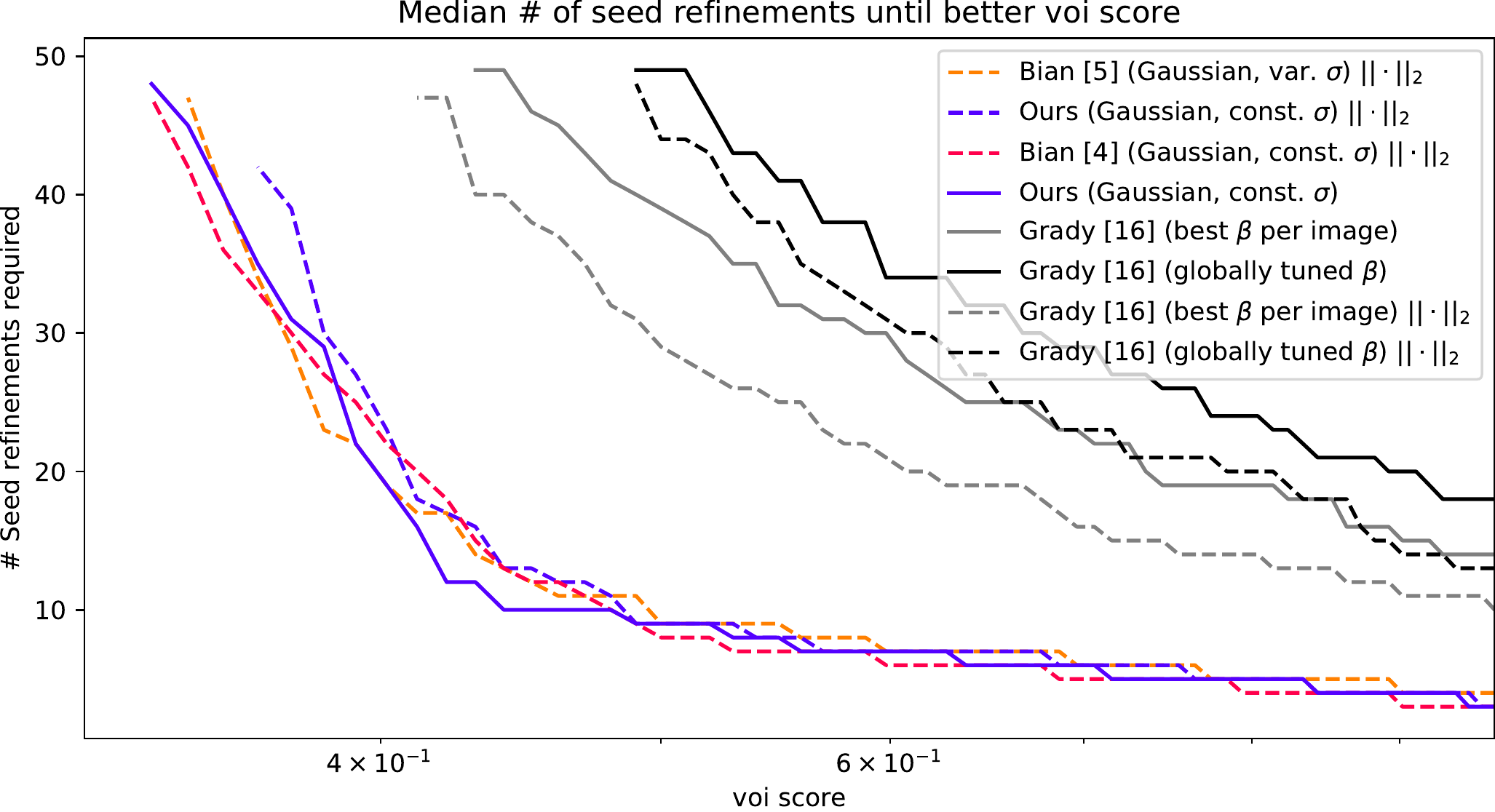}
    \includegraphics[width=0.49\columnwidth]{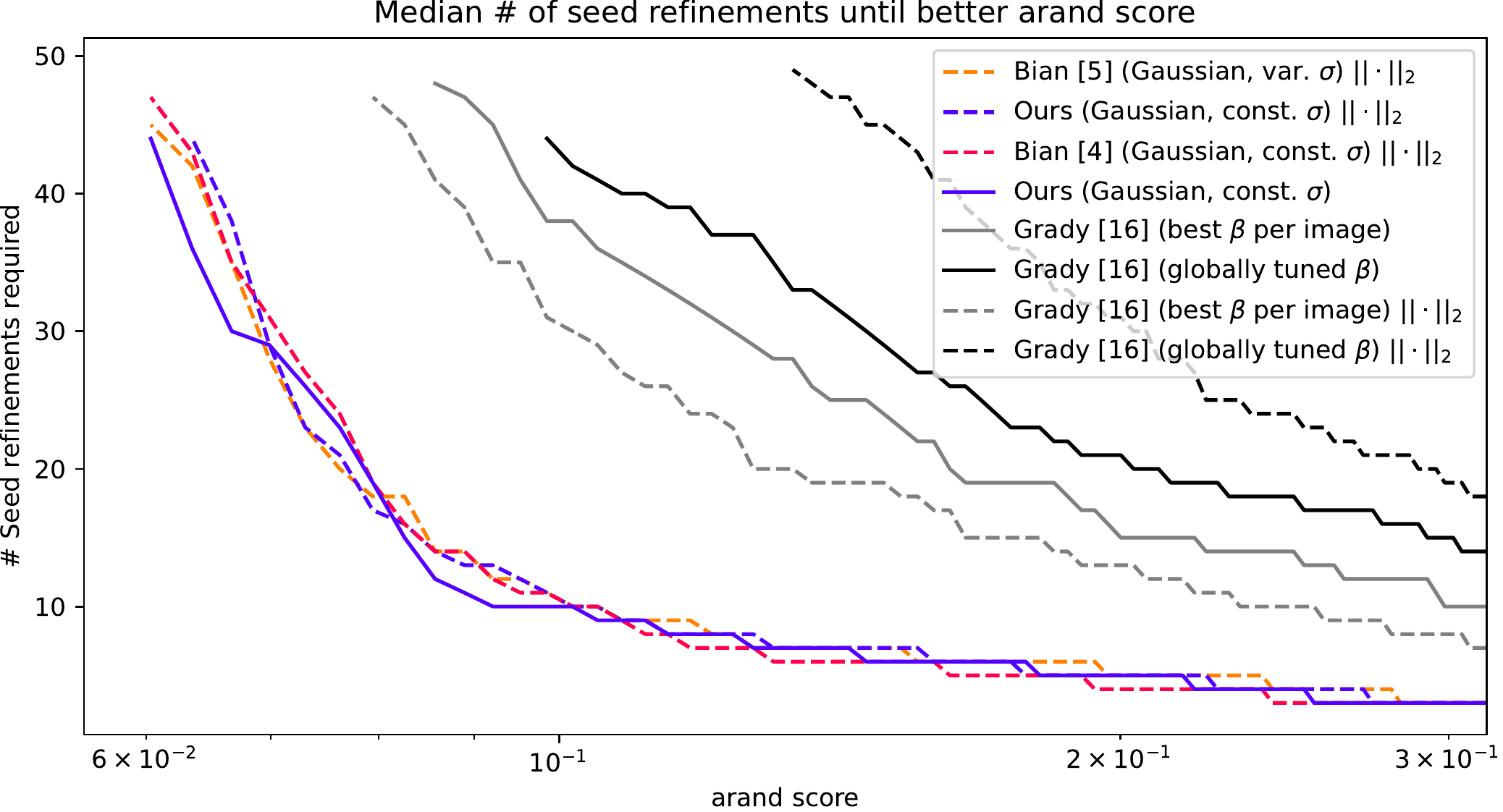}
    \caption{fastMRI dataset experiments:
       Mean VOI/ARAND scores after placing $n$ additional seeds (top) and median number of additional seeds required to achieve a specific VOI/ARAND score (bottom). Seeds were set as explained in \autoref{subsec:eval_meth}.
    }\label{fig:eval_quantitative_mri}
\end{figure}

\vspace{1mm}
\noindent
{\bf FastMRI dataset.}
Image data for this dataset were obtained from the NYU fastMRI Initiative database \cite{knoll2020fastmri,zbontar2018fastmri} (publicly available at: fastmri.med.nyu.edu), where a listing of NYU fastMRI investigators, which provided data but did not contribute to the work in any other way, can also be found.
The complex valued k-space single coil knee data was converted into the complex image space by inverse Fourier transform.
100 images were hand labeled into four classes: upper bone, lower bone, knee tissue, and background. The masks are available online\footnote{\url{https://uni-muenster.sciebo.de/s/DK9F0f6p5ppsWXC}}. 
The resulting complex image (isomorphic to 2D vector image) can be processed as-is by Grady's method and the proposed weight function for the multivariate Gaussian case.
Bian's methods, that only operate on scalar value images, are applied to the magnitude images.

Overall, the mean scores imply that in this specific case the proposed method for the appropriate image model (2D Gaussian noise) performs roughly on par with competing methods by Bian et al.~ \cite{ang2016statistical,ang2016ttest} operating on the magnitude image (\autoref{fig:eval_quantitative_mri}), which suggests that the practice of approximating the Rician distribution in MRI magnitude images with a Gaussian~\cite{Aja-Fernandez2016noiseMRI} works well in practice.
Notably, our method benefits from the 2D information compared to the magnitude-variant, while Grady's method benefits only in some cases and performs considerably worse overall.

\section{Conclusion}
We have presented a general framework to derive noise model specific weight functions for random walker segmentation. 
Under the assumption of a known noise model, our framework enables the derivation of a weight function based on the Bhattacharyya coefficient, which takes the pixel distribution into account and is thus robust against noise.
We have derived the specific weight functions for Poisson and Gaussian noise with global and region-specific variance and show their suitability in the segmentation of synthetic as well as real world data.

Our method may further be applied to other domains by computing the explicit weight functions in \autoref{eq:framework}. Examples include magnitude images with Rician noise~\cite{Aja-Fernandez2016noiseMRI}, raw images of digital imaging sensors with Poissonian-Gaussian noise~\cite{foi2008practical}, SAR images with speckle noise modeled by Gamma, Weibull or Fisher distributions~\cite{DBLP:journals/tgrs/LuoSTG20a,DBLP:journals/tgrs/WangWLZZ17}.
Additionally, extensions of the random walker algorithm (e.g. random walker with restarts \cite{Ham2013,kim2008generative}, non-local random walker \cite{improvingrw2022wang}) can easily be incorporated into our framework. Applications beyond segmentation could also benefit from the improved weight function, such as random walker for visual tracking \cite{li2015visual}, target detection \cite{qin2019infrared}, and saliency detection \cite{Jian2018}.

Since the main contribution of this work is the definition and derivation of a similarity measure for pixel values in the presence of noise, this similarity measure and the related Hellinger distance function \cite{hellinger1909neue} can be used in other methods, e.g. graph cut algorithm \cite{boykov2000interactive}, live wire segmentation \cite{mortensen1992adaptive} and extensions. 

\section*{Acknowledgements}
We thank qubeto GmbH and Julian Bigge for providing the FIM larvae images and Jiaqi Zhang for her support with data annotations.

\bibliographystyle{splncs04}
\bibliography{src}

\clearpage

\appendix

{
\noindent\Large\textbf{Appendix}
}
\vspace{5mm} \\
This Appendix briefly describes the code of the python library (\autoref{supsec:library}) and data used for evaluation (\autoref{supsec:data}) before providing a more detailed version of \autoref{sec:noise_models} in \autoref{supsec:noise_models} of the appendix. All contents presented here, i.e. the python library, the Data including annotations and the detailed derivations, are publicly available.

\section{Python Library}
\label{supsec:library}
We provide the python library that implements the random walker method in combination with the presented weight functions (in addition to the weight functions by Grady~\cite{grady2006random} and Bian et al.~\cite{ang2016statistical,ang2016ttest}) online\footnote{\url{https://zivgitlab.uni-muenster.de/ag-pria/rw-noise-model}}.
In addition to the library code itself, all scripts used to generate quantitative and qualitative results in the paper are also provided in the sub folder \texttt{evaluation}.
Please refer to the provided \texttt{Readme.md} files in the repository for details about compilation and usage.

\section{Data Used for Evaluation}
\label{supsec:data}

In order to facilitate reproducibility of the reported results, we also make the data that was used for quantitative and qualitative evaluation available online\footnote{\url{https://uni-muenster.sciebo.de/s/DK9F0f6p5ppsWXC}}.
This includes the FIM~\cite{risse2013fim} larvae images with corresponding pixel-wise ground truth class labels and the pixel-wise class labels for images from the FastMRI~\cite{knoll2020fastmri} dataset.
Due to restrictions imposed by the creators of the FastMRI dataset, we are unable redistribute the reconstructed MRI files directly, but we provide detailed a description of how to create the used images in the \texttt{Readme.md} file in the \texttt{evaluation} sub folder in the code repository (see above).

\section{Detailed Derivations for Concrete Noise Models}
\label{supsec:noise_models}

In this section, we derive all the equations in more detail.
For the sake of readability and to avoid having to jump between the two documents while reading, we provide all content from \autoref{sec:noise_models} from the main paper here again, but also reference the main paper when necessary.
The equations are numbered continuously, which means that equations (1)-(10) are in the main paper and this Appendix starts from equation (11).

In this section we apply the framework presented in \autoref{sec:framework} of the main paper to three noise models relevant to various imaging modalities and obtain closed form solutions.
We choose Poisson noise as it is very common in biology \cite{risse2013fim,micronoise2006Vonesch}, as well as Gaussian noise - with two different configurations: fixed variance over the whole image and variable variance per image region - which is common in many images including medically relevant techniques like MRI and CT.
For Poisson noise this is the first derivation of a noise model specific weight function for random walker, whereas the two versions of Gaussian noise model have been considered in \cite{ang2016statistical,ang2016ttest}, respectively.

\subsection{Poisson Noise}\label{sec:A_poisson}
In this subsection we assume an image that is affected by Poisson noise (also called ``shot noise``), which occurs, for example, as an effect of photon or electron counting in imaging systems (e.g., fluorescence microscopy \cite{micronoise2006Vonesch}, positron emission tomography \cite{petnoise1982Shepp}, low dose CT \cite{ctnoise2001Lu}).
In this model the measured pixel values $x,y \in \mathbb{N}$ are drawn from a Poisson distribution with unknown parameter $\kappa = \lambda$, which corresponds to the true pixel value.
We assume the prior distribution of $\lambda$ to be uniform in $P=(0, a)$ for some sufficiently large value $a$.
For convenience, we define $\sums{X}=\sum_{x \in X} x$ and $\sums{Y}=\sum_{y \in Y} y$. The following calculations give a detailed version of \autoref{eq:poisson_gamma} from the main paper.
\begin{linenomath}
\begin{align}
    w_{\mathcal{X}\mathcal{Y}}=BC(p(\cdot | X), p(\cdot | Y)) = &\frac{\int_{P_\kappa} \sqrt{\prod_{x \in X} p(x | \kappa) \prod_{y \in Y}p(y | \kappa)}d\kappa}{\sqrt{\int_{P_\kappa} \prod_{x \in X} p(x | \kappa) d\kappa \int_{P_\kappa} \prod_{y \in Y} p(y | \kappa) d\kappa}}\\
    = &\frac{\sqrt{\int_P \prod_{x \in X} \frac{e^{-\lambda}\lambda^{x}}{x!} \prod_{y \in Y} \frac{e^{-\lambda}\lambda^{y}}{y!}}d\lambda}{\sqrt{\int_P \prod_{x \in X} \frac{e^{-\lambda}\lambda^{x}}{x!} d\lambda \int_P \prod_{y \in Y} \frac{e^{-\lambda}\lambda^{y}}{y!} d\lambda}}\\
    = &\frac{\int_P \sqrt{ \prod_{x \in X} e^{-\lambda}\lambda^{x}e^{-\lambda} \prod_{y \in Y} \lambda^{y}}d\lambda}{\sqrt{\int_P \prod_{x \in X} e^{-\lambda}\lambda^{x} d\lambda \int_P \prod_{y \in Y} e^{-\lambda}\lambda^{y} d\lambda}}\\
    = &\frac{\int_P  \sqrt{e^{-n\lambda}\lambda^{\sum_{x \in X} x}e^{-n\lambda}\lambda^{\sum_{y \in Y} y_i}}d\lambda}{\sqrt{\int_P  e^{-n\lambda}\lambda^{\sum_{x \in X} x} d\lambda \int_P e^{-n\lambda}\lambda^{\sum_{y \in Y} y} d\lambda}}\\
    = &\frac{\int_P  e^{-n\lambda} \lambda^{\frac{\sums{X}+\sums{Y}}{2}}d\lambda}{\sqrt{\int_P  e^{-n\lambda}\lambda^{\sums{X}} d\lambda \int_P e^{-n\lambda}\lambda^{\sums{Y}} d\lambda}}\\
    = &\frac{\int_P  e^{-n\lambda} \lambda^{\frac{\sums{X}+\sums{Y}}{2}}d\lambda}{\sqrt{\int_P  e^{-n\lambda}\lambda^{\sums{X}} d\lambda \int_P e^{-n\lambda}\lambda^{\sums{Y}} d\lambda}} \frac{n^{\frac{\sums{X}+\sums{Y}}{2}}}{\sqrt{n^X n^{\sums{Y}}}}\\
    = &\frac{\int_{P_a}  e^{-n\lambda} (n\lambda)^{\frac{\sums{X}+\sums{Y}}{2}}d\lambda}{\sqrt{\int_{P_a}  e^{-n\lambda}(n\lambda)^{\sums{X}} d\lambda \int_{P_a} e^{-n\lambda}(n\lambda)^{\sums{Y}} d\lambda}}\label{eq:poi_eas}
\end{align}
\end{linenomath}
\begin{linenomath}
\begin{align}
    \overset{L:= n\lambda}{=} &\frac{\int_{P_{na}}  e^{-L} L^{\frac{\sums{X}+\sums{Y}}{2}}\frac{1}{n}dL}{\sqrt{\int_{P_{na}}  e^{-L}L^{\sums{X}} \frac{1}{n} dL \int_{P_{na}} e^{-L}L^{\sums{Y}} \frac{1}{n} dL}}\\
    \underset{a \rightarrow \infty}{\rightarrow} &\frac{\int_0^{\infty}  e^{-L} L^{\frac{\sums{X}+\sums{Y}}{2}}dL}{\sqrt{\int_0^{\infty}  e^{-L}L^{\sums{X}} dL \int_0^{\infty} e^{-L}L^{\sums{Y}} dL}}\label{subeq:conv}\\
    = &\frac{\Gamma(\frac{\sums{X}+\sums{Y}}{2} + 1)}{\sqrt{\Gamma(\sums{X}+1)\Gamma(\sums{Y}+1)}}\label{subeq:poisson_gamma}
\end{align}
\end{linenomath}
After inserting the definition of the Poisson PDF into \autoref{eq:framework} simple calculations lead to the equation in \autoref{eq:poi_eas}.
Then a substitution of $L:=n\lambda$ is used.
Afterwards we let $a$ tend to infinity (see below for discussion on the convergence) and in the last line, the (integral) definition of the Gamma function is used.

About the convergence in \autoref{subeq:conv}: Since we cannot a priori set $ a = \infty$, because there is no uniform distribution on $\mathbb{R}$, the resulting formula in \autoref{eq:poisson_gamma} does not equal the fraction of Gamma functions, but some approximation of that.
Since however, the exponential function $e^{-L}$ decreases very fast, even in comparison to the polynomial term $L^{\sums{X}}$ (and according polynomial terms in the other integrals), the precision for large enough $a$ is good enough to assume equality in numerical computations.

In practice, the \texttt{lgamma} function (i.e., the logarithm of gamma function, provided in standard python and c++ libraries) should be used to avoid numerical overflows:
\begin{linenomath}
\begin{align}
    w_{\pixel{X}\pixel{Y}} = \exp\left(\text{lgamma}\left({\frac{\sums{X}+\sums{Y}}{2}+1}\right) - \frac{\text{lgamma}(\sums{X}+1)+\text{lgamma}(\sums{Y}+1)}{2}\right)
\end{align}
\end{linenomath}
If this function is unavailable, one can also use the approximation $w_{\mathcal{X}\mathcal{Y}}=\exp(-\frac{1}{2} (\sqrt{\sums{X}} - \sqrt{\sums{Y}})^2)$,
which is the Bhattacharyya-coefficient for two Poisson probability density functions~\cite{nielsen2011burbea} with $\lambda_1 = \sums{X}$ and $\lambda_2 = \sums{Y}$, respectively.
Experiments suggest that the absolute difference between this approximation and \autoref{subeq:poisson_gamma} is below 0.05 for $\sums{X},\sums{Y} > 2$.

\subsection{Multivariate Gaussian Noise with Constant Covariance}

In this subsection we assume an m-channel image with concrete pixel values $x,y \in \mathbb{R}^m$.
The true image values are perturbed by additive Gaussian noise and are thus modeled by a Gaussian PDF with (unknown) parameter $\kappa = \mu \in \mathbb{R}^m$, which also corresponds to the true pixel value.
This noise model applies, for example, in complex valued MRI images \cite{mrinoise1985henkelman}.
Further, we assume that $\mu$ is priorly uniformly distributed in $P=(-a, a)^m$ for some sufficiently large value $a$.
The covariance matrix $C$ is assumed to be constant for the whole image.
It should be noted that the special case $m=1$ is the setting that is assumed in prior work~\cite{ang2016statistical}.
Starting from \autoref{eq:param_given_set} we obtain:

\begin{linenomath}
\begin{align}
    p(\mu | X) = &\frac{\prod_{x \in X} p(x | \mu)} {\int_P \prod_{x \in X} p(x | \tilde{\mu}) d\tilde{\mu}}\\
                = &\frac{\prod_{x \in X} \frac{1}{\sqrt{(2\pi)^m \det(C)}} \exp(-\frac{1}{2} (x - \mu)^T C^{-1} (x - \mu))} {\int_P \prod_{x \in X} \frac{1}{\sqrt{(2\pi)^m \det(C)}} \exp(-\frac{1}{2} (x - \tilde{\mu})^T C^{-1} (x - \tilde{\mu})) d\tilde{\mu}}\\
                = &\frac{\prod_{x \in X} \exp(-\frac{1}{2} (x - \mu)^T C^{-1} (x - \mu))} {\int_P \prod_{x \in X} \exp(-\frac{1}{2} (x - \tilde{\mu})^T C^{-1} (x - \tilde{\mu})) d\tilde{\mu}}\\
                = &\frac{\exp(-\frac{1}{2} \sum_{x \in X} (x - \mu)^T C^{-1} (x - \mu))} {\int_P \exp(-\frac{1}{2} \sum_{x \in X} (x - \tilde{\mu})^T C^{-1} (x - \tilde{\mu})) d\tilde{\mu}}\label{subeq:mdim_gauss_pdf1}
\end{align}
\end{linenomath}

We consider the sum term in the exponents separately first:
\begin{linenomath}
\begin{align}
      &\sum_{x\in X} (x - \mu)^T C^{-1} (x - \mu)\nonumber\\
    = &\sum_{x\in X} x^T C^{-1} x - \sum_{x\in X} x^T C^{-1} \mu - \sum_{x\in X} \mu^T C^{-1} x + n \mu^T C^{-1} \mu\\
    = &\sum_{x\in X} x^T C^{-1} x - n\left(\sum_{x\in X} \frac{x^T}{n} C^{-1} \mu - \sum_{x\in X} \mu^T C^{-1} \frac{x}{n} + \mu^T C^{-1} \mu\right)\\
    = &\sum_{x\in X} x^T C^{-1} x - n(\sum_{x\in X} \frac{x}{n})^T C^{-1} (\sum_{x\in X} \frac{x}{n}) + n(\sum_{x\in X} \frac{x}{n})^T C^{-1} (\sum_{x\in X} \frac{x}{n}) \\
      & - n\left(\sum_{x\in X} \frac{x}{n}^T C^{-1} \mu - \sum_{x\in X} \mu^T C^{-1} \frac{x}{n} + \mu^T C^{-1} \mu\right)\\
    = &\sum_{x\in X} x^T C^{-1} x - \frac{1}{n}(\sum_{x\in X} x)^T C^{-1} (\sum_{x\in X} x) + (\sum_{x\in X} \frac{x}{n} - \mu)^T C^{-1} (\sum_{x\in X} \frac{x}{n} - \mu)) \label{eq:before_os}\\
    =: & O_S + (\sum_{x\in X} \frac{x}{n} - \mu)^T \left(\frac{C}{n}\right)^{-1} (\sum_{x\in X} \frac{x}{n} - \mu)\label{eq:exp_sum1}
\end{align}
\end{linenomath}

It should be noted that the first two summands of \autoref{eq:before_os} is defined as $O_S$, which is independent of $\mu$. Inserting this into \autoref{subeq:mdim_gauss_pdf1} we can simplify further:

\begin{linenomath}
\begin{align}
    p(\mu | X) = &\frac{\exp(-\frac{1}{2} \sum_{x\in X} (x - \mu)^T C^{-1} (x - \mu))} {\int_P \exp(-\frac{1}{2} \sum_{x\in X} (x - \tilde{\mu})^T C^{-1} (x - \tilde{\mu})) d\tilde{\mu}}\\
                = &\frac{\exp(-\frac{1}{2} (O_S + (\sum_{x\in X} \frac{x}{n} - \mu)^T \left(\frac{C}{n}\right)^{-1} (\sum_{x\in X} \frac{x}{n} - \mu))} {\int_P \exp(-\frac{1}{2} (O_S + (\sum_{x\in X} \frac{x}{n} - \tilde{\mu})^T \left(\frac{C}{n}\right)^{-1} (\sum_{x\in X} \frac{x}{n} - \tilde{\mu})) d\tilde{\mu}}\\
                = &\frac{\exp(-\frac{1}{2} O_S ) \exp(-\frac{1}{2} (\sum_{x\in X} \frac{x}{n} - \mu)^T \left(\frac{C}{n}\right)^{-1} (\sum_{x\in X} \frac{x}{n} - \mu))} {\int_P \exp(-\frac{1}{2} O_S) \exp(-\frac{1}{2} (\sum_{x\in X} \frac{x}{n} - \tilde{\mu})^T \left(\frac{C}{n}\right)^{-1} (\sum_{x\in X} \frac{x}{n} - \tilde{\mu})) d\tilde{\mu}}\\
                = &\frac{\exp(-\frac{1}{2} (\sum_{x\in X} \frac{x}{n} - \mu)^T \left(\frac{C}{n}\right)^{-1} (\sum_{x\in X} \frac{x}{n} - \mu))} {\int_P \exp(-\frac{1}{2} (\sum_{x\in X} \frac{x}{n} - \tilde{\mu})^T \left(\frac{C}{n}\right)^{-1} (\sum_{x\in X} \frac{x}{n} - \tilde{\mu})) d\tilde{\mu}}\\
                = &\frac{\frac{1}{\sqrt{(2\pi)^m \det(\frac{C}{n})}} \exp(-\frac{1}{2} (\sum_{x\in X} \frac{x}{n} - \mu)^T \left(\frac{C}{n}\right)^{-1} (\sum_{x\in X} \frac{x}{n} - \mu))} {\int_P \frac{1}{\sqrt{(2\pi)^m \det(\frac{C}{n})}} \exp(-\frac{1}{2} (\sum_{x\in X} \frac{x}{n} - \tilde{\mu})^T \left(\frac{C}{n}\right)^{-1} (\sum_{x\in X} \frac{x}{n} - \tilde{\mu})) d\tilde{\mu}}\\
	\stackrel{a\to \infty}{\to} &\frac{\frac{1}{\sqrt{(2\pi)^m \det(\frac{C}{n})}} \exp(-\frac{1}{2} (\sum_{x\in X} \frac{x}{n} - \mu)^T \left(\frac{C}{n}\right)^{-1} (\sum_{x\in X} \frac{x}{n} - \mu))} {\int_{\mathbb{R}^m} \frac{1}{\sqrt{(2\pi)^m \det(\frac{C}{n})}} \exp(-\frac{1}{2} (\sum_{x\in X} \frac{x}{n} - \tilde{\mu})^T \left(\frac{C}{n}\right)^{-1} (\sum_{x\in X} \frac{x}{n} - \tilde{\mu})) d\tilde{\mu}}\label{subeq:conv2}\\
                 =&\frac{1}{\sqrt{(2\pi)^m \det(\frac{C}{n})}} \exp\left(-\frac{1}{2} (\mu - \sum_{x\in X} \frac{x}{n})^T \left(\frac{C}{n}\right)^{-1} (\mu - \sum_{x\in X} \frac{x}{n})\right)\label{subeq:mdim_gauss_pdf2}
\end{align}
\end{linenomath}
In the last step, we make use of the fact that the denominator is the integral over the density of a Gaussian distribution, which is 1.
 About the convergence in \autoref{subeq:conv2}: Since we can not a priori set $ a = \infty$, because there is no uniform distribution on $\mathbb{R}^m$, the resulting distribution will not exactly be Gaussian, but Gaussian, where the density is "cut off" in all directions at $-a$ and $a$ (and scaled accordingly).
 Since we have that this density function converges point wise to the density function of a Gaussian distribution, we also have Convergence in Distribution for the according random variables (with Scheffés Lemma).

 \autoref{subeq:mdim_gauss_pdf2} shows that $p(\mu | X)$ is simply the density function of a normal distribution with mean $\sum_{x\in X} \frac{x}{n} =: \bar{X}$ and covariance matrix $\frac{C}{n}$ (and $p(\mu|Y)$ accordingly).
 As shown in \cite{nielsen2011burbea} the Bhattacharyya coefficient is then:
\begin{linenomath}
\begin{align}\label{supeq:mult_gauss_fin}
	w_{\mathcal{X}\mathcal{Y}} & = BC(p(\cdot | X), p(\cdot | Y)) \\
	& = \exp\bigg(-\frac{1}{8}(\bar{X}- \bar{Y})^T\left(\frac{C}{n}\right)^{-1}(\bar{X} - \bar{Y})\bigg)
\end{align}
\end{linenomath}
It should be noted that in \cite{nielsen2011burbea} a square root is missing in the denominator of the second summand of the respective formula in Table 1.

\subsection{Gaussian Noise with Signal-Dependent Variance}

In this subsection we assume additive Gaussian noise on single channel images where, however, $\sigma^2$ differs between image regions (in contrast to \autoref{sec:gauss} in the main paper which assumes a global, constant $C=\sigma^2$ for m=1).
Thus, pixel values are modeled by $x = \mu_\pixel{X} + \mathcal{N}(0, \sigma_\pixel{X}^2)$.
A special case is Loupas noise, where $\sigma_\pixel{X}^2 = \sqrt{\mu_\pixel{X}}\sigma^2$ for some fixed (global) $\sigma^2$. It applies, for example, to speckled SAR and medical ultrasound images~\cite{ang2016ttest,TENBRINCK2015arbnoise}
Thus, we have to estimate $\mu$ and $\sigma^2$ simultaneously and set $\kappa$ from \autoref{eq:framework} to be $(\mu, \sigma^2)$.

Before go into solving \autoref{eq:framework} for this case, we calculate the general form of an integral that will be used later:
\begin{linenomath}
\begin{align}\label{eq:int_int}
    & \left(\frac{1}{2\pi}\right)^{m-1} \int_{(0,\infty)} y^{\frac{-(m-1)}{2}} \exp \left(-\frac{1}{y}\frac{1}{4m}\sum_{i,j=1}^m (x_i - x_j)^2 \right) dy \\ \label{eq:int_sub}
   = &  \left(\frac{1}{2\pi}\right)^{m-1} \int_{(0,\infty)} y^{\frac{3}{2}(m-1)} \left| -3y^{-4} \right| \exp \left(-y^3 \frac{1}{4m}\sum_{i,j=1}^m (x_i - x_j)^2 \right) dy\\
   = & 3 \left(\frac{1}{2\pi}\right)^{m-1} \int_{(0,\infty)} y^{\frac{3m-11}{2}} \exp \left(-y^3 \frac{1}{4m}\sum_{i,j=1}^m (x_i - x_j)^2 \right) dy\\\label{eq:int_wiki}
   = & 3 \left(\frac{1}{2\pi}\right)^{m-1} \frac{1}{3} \left(\frac{1}{4m}\sum_{i,j=1}^m (x_i - x_j)^2 \right)^{\frac{-m+3}{2}}\Gamma\left(\frac{m-3}{2}\right) \\ \label{eq:int_int_end}
   = & \left(\frac{1}{2\pi}\right)^{m-1} \Gamma\left(\frac{m-3}{2}\right)\left(\frac{1}{4m}\sum_{i,j=1}^m (x_i - x_j)^2 \right)^{\frac{-m+3}{2}}
\end{align}
\end{linenomath}
We used:
\begin{enumerate}
    \item In \autoref{eq:int_sub}: Substitution of integration variable with $y\mapsto y^{-3}$, with substitution derivative $-3y^{-4}$.
    \item In \autoref{eq:int_wiki}: Formula from the collection of integrals, series and products \cite{gradshteyn2014table}, where $m\geq4$ is required, which does not matter in practice, since $m$ will be the size of the neighborhood, where the smallest realistic size is $3 \times 3=9$.
\end{enumerate}
To solve \autoref{eq:framework} we set $P_a := (0, a)\times(-a,a)$ (again, $a$ sufficiently large) and assume the prior distribution of $(\mu, \sigma^2)$ to be uniform.
We then have to calculate the integrals in the enumerator and denominator of \autoref{eq:framework}, which can be reformulated to a similar form as \autoref{eq:int_int} and then can be calculated analogously.
It should be noted that in the following derivation, the terms from the third row on span two lines each.
\begin{linenomath}
\begin{align}
   & BC(p(\cdot | X), p(\cdot | Y)) \nonumber \\
   & = \frac{\int_P \prod_{x\in X} \sqrt{p(x | (\mu, \sigma^2))p(y | (\mu, \sigma^2))}d(\mu,\sigma^2)}{\sqrt{\int_P \prod_{x\in X} p(x | (\mu, \sigma^2)) d(\mu,\sigma^2) \int_P \prod_{x\in X} p(y | (\mu, \sigma^2)) d(\mu,\sigma^2)}}\\
   & = \frac{\int_P \prod_{z\in X\cup Y} \sqrt{p(z | (\mu, \sigma^2))}d(\mu,\sigma^2)}{\sqrt{\int_P \prod_{x\in X} p(x | (\mu, \sigma^2)) d(\mu,\sigma^2) \int_P \prod_{y\in Y} p(y | (\mu, \sigma^2)) d(\mu,\sigma^2)}}\\
   & = \frac{\int_P \prod_{z\in X\cup Y} \sqrt{\frac{1}{\sqrt{2\pi \sigma^2}}\exp(-\frac{1}{2\sigma^2}(z - \mu)^2)}d(\mu,\sigma^2)}{\sqrt{\int_P \prod_{x\in X} \frac{1}{\sqrt{2\pi \sigma^2}}\exp(-\frac{1}{2\sigma^2}(x - \mu)^2)) d(\mu,\sigma^2)}} \nonumber \\
   & \frac{1}{\sqrt{\int_P \prod_{y\in Y} \frac{1}{\sqrt{2\pi \sigma^2}}\exp(-\frac{1}{2\sigma^2}(y - \mu)^2))  d(\mu,\sigma^2)}}\\
   & = \frac{\int_P \frac{1}{\sqrt{2\pi \sigma^2}}^{n}\sqrt{\exp(-\frac{1}{2\sigma^2}\sum_{z\in X\cup Y} (z - \mu)^2)}d(\mu,\sigma^2)}{\sqrt{\int_P \frac{1}{\sqrt{2\pi \sigma^2}}^n\exp(-\frac{1}{2\sigma^2}\sum_{x\in X} (x - \mu)^2) d(\mu,\sigma^2)}} \nonumber\\
   & \frac{1}{\sqrt{\int_P \frac{1}{\sqrt{2\pi \sigma^2}}^n\exp(-\frac{1}{2\sigma^2}\sum_{y\in Y} (y - \mu)^2)  d(\mu,\sigma^2)}}\\
   & = \frac{\int_P \frac{1}{\sqrt{2\pi \sigma^2}}^{n}\exp(-\frac{1}{4\sigma^2}\sum_{z\in X\cup Y} (z - \mu)^2)d(\mu,\sigma^2)}{\sqrt{\int_P \frac{1}{\sqrt{2\pi \sigma^2}}^n\exp(-\frac{1}{2\sigma^2}\sum_{x\in X} (x - \mu)^2)) d(\mu,\sigma^2)}} \nonumber \\
   & \frac{1}{\sqrt{\int_P \frac{1}{\sqrt{2\pi \sigma^2}}^n\exp(-\frac{1}{2\sigma^2}\sum_{y\in Y} (y - \mu)^2)) d(\mu,\sigma^2)}}
   \label{eq:bc_sigma}
\end{align}
\end{linenomath}

We again first calculate the exponent, to get to the form of \autoref{eq:int_int}. The first equation follows in similar fashion as \autoref{eq:exp_sum1}.
\begin{linenomath}
\begin{align}
	\sum_{x_1\in X} (x_1 - \mu)^2 &= n (\frac{1}{n}\sum_{x_1\in X} x_1 - \mu)^2 + \sum_{x_1\in X} x_1^2 - \frac{1}{n}(\sum_{x_1\in X} x_1)^2\\
                                &= n (\frac{1}{n}\sum_{x_1\in X} x_1 - \mu)^2 + \sum_{x_1\in X} x_1^2 - \frac{1}{n}\sum_{x_1\in X} \sum_{x_2\in X} x_1 x_2\\
                                &= n (\frac{1}{n}\sum_{x_1\in X} x_1 - \mu)^2 + \frac{1}{2n}(n\sum_{x_1\in X} x_1^2 + n\sum_{x_2\in X} x_2^2 - \sum_{x_1\in X} \sum_{x_2\in X} 2 x_1 x_2 )
\end{align}
\end{linenomath}
\begin{linenomath}
\begin{align}
                                &= n (\frac{1}{n}\sum_{x_1\in X} x_1 - \mu)^2 + \frac{1}{2n}(\sum_{x_1\in X} \sum_{x_2\in X} x_1^2 + x_2^2 - 2 x_1 x_2 )\\
                                &= n (\frac{1}{n}\sum_{x_1\in X} x_1 - \mu)^2 + \frac{1}{2n} \sum_{x_1\in X} \sum_{x_2\in X} (x_1 - x_2)^2 \label{eq:solving_sum}
\end{align}
\end{linenomath}

Using \autoref{eq:solving_sum} we can further simplify \autoref{eq:bc_sigma}. We first advance with one factor of the denominator of equation \autoref{eq:bc_sigma}:
\begin{linenomath}
\begin{align}
    & \int_P \frac{1}{\sqrt{2\pi \sigma^2}}^n\exp\left(-\frac{1}{2\sigma^2}\sum_{x_1\in X} (x_1 - \mu)^2\right) d(\mu,\sigma^2) \nonumber \\
    = & \int_P \frac{1}{\sqrt{2\pi \sigma^2}}^n\exp\left(-\frac{n}{2\sigma^2}(\frac{1}{n}\sum_{x_1\in X} x_1 - \mu)^2 - \frac{1}{2\sigma^2} \frac{1}{2n} \sum_{x_1\in X} \sum_{x_2\in X} (x_1 - x_2)^2\right) d(\mu,\sigma^2) \\
    = & \int_{(0,a)} \frac{1}{\sqrt{2\pi \sigma^2}}^{n-1} \exp \left(-\frac{1}{4\sigma^2n} \sum_{x_1\in X} \sum_{x_2\in X} (x_1 - x_2)^2\right) \nonumber\\
      & \int_{(-a, a)} \frac{1}{\sqrt{2\pi \sigma^2}} \exp \left( -\frac{n}{2\sigma^2}\left(  \frac{1}{n}\sum_{x_1\in X} x_1 - \mu\right)^2 \right) d\mu \ d\sigma^2 \\
    = & \int_{(0,a)} \frac{1}{\sqrt{2\pi \sigma^2}}^{n-1} \exp \left(-\frac{1}{4\sigma^2n} \sum_{x_1\in X} \sum_{x_2\in X} (x_1 - x_2)^2\right) \nonumber\\
      & \int_{(-a, a)} \frac{1}{\sqrt{n}}\frac{1}{\sqrt{2\pi \frac{\sigma^2}{n}}} \exp \left( -\frac{1}{2\frac{\sigma^2}{n}}\left(  \frac{1}{n}\sum_{x_1\in X} x_1 - \mu\right)^2 \right) d\mu \ d\sigma^2 \\
    \underset{a \rightarrow \infty}{\rightarrow} & \int_{(0,\infty)} \frac{1}{\sqrt{2\pi \sigma^2}}^{n-1} \exp \left(-\frac{1}{4\sigma^2n} \sum_{x_1\in X} \sum_{x_2\in X} (x_1 - x_2)^2\right) \nonumber\\
      & \frac{1}{\sqrt{n}}\int_{(-\infty, \infty)} \frac{1}{\sqrt{2\pi \frac{\sigma^2}{n}}} \exp \left( -\frac{1}{2\frac{\sigma^2}{n}}\left(  \frac{1}{n}\sum_{x_1\in X} x_1 - \mu\right)^2 \right) d\mu \ d\sigma^2 \\  \label{eq:normal_density}
    = & \frac{1}{\sqrt{n}} \int_{(0,\infty)} \frac{1}{\sqrt{2\pi \sigma^2}}^{n-1} \exp \left(-\frac{1}{4\sigma^2n} \sum_{x_1\in X} \sum_{x_2\in X} (x_1 - x_2)^2\right) d\sigma^2\\ \label{eq:nom_int_last}
    = & \frac{1}{\sqrt{n}} \left(\frac{1}{2\pi}\right)^{n-1} \Gamma\left(\frac{n-3}{2}\right)\left(\frac{1}{4n}\sum_{x_1, x_2 \in X} (x_1 - x_2)^2 \right)^{\frac{-n+3}{2}}
\end{align}
\end{linenomath}

Here we used that the inner Integral in \autoref{eq:normal_density} is the density function of a Gaussian distribution with Variance $\frac{\sigma^2}{n}$ and expected value $\frac{1}{n}\sum_{i=1}^n x_i$ (note, we are integrating by $\mu$), which integrates to 1. In \autoref{eq:nom_int_last} we make use of \autoref{eq:int_int_end}.

The last remaining step is to apply equations leading to \autoref{eq:nom_int_last} to the nominator and denominator in \autoref{eq:bc_sigma}. We get
\begin{linenomath}
\begin{align}
    & \int_P \frac{1}{\sqrt{2\pi \sigma^2}}^n\exp\left(-\frac{1}{2\sigma^2}\sum_{x \in X} (x - \mu)^2)\right) d(\mu,\sigma^2) \nonumber \\
    = & \frac{1}{\sqrt{n}} \left(\frac{1}{2\pi}\right)^{n-1} \Gamma\left(\frac{n-3}{2}\right)\left(\frac{1}{4n}\sum_{x_1, x_2 \in X}^n (x_1 - x_2)^2 \right)^{\frac{-n+3}{2}} \ \ \ \text{and} \\
    & \int_P \frac{1}{\sqrt{2\pi \sigma^2}}^n\exp\left(-\frac{1}{4\sigma^2}\sum_{z \in X\cup Y} (z - \mu)^2)\right) d(\mu,\sigma^2) \nonumber \\
    = & \frac{1}{\sqrt{n}} \left(\frac{1}{2\pi}\right)^{n-1} \Gamma\left(\frac{n-3}{2}\right)\left(\frac{1}{16n}\sum_{z_1, z_2 \in X\cup Y} (z_1 - z_2)^2 \right)^{\frac{-n+3}{2}}
\end{align}
The full fraction then reads:
\begin{align}
    \frac{\frac{1}{\sqrt{n}} \left(\frac{1}{2\pi}\right)^{n-1} \Gamma\left(\frac{n-3}{2}\right)\left(\frac{1}{16n}\sum_{z_1, z_2 \in X\cup Y} (z_1 - z_2)^2 \right)^{\frac{-n+3}{2}}}{\sqrt{\left(\frac{1}{\sqrt{n}} \left(\frac{1}{2\pi}\right)^{n-1} \Gamma\left(\frac{n-3}{2}\right)\right)^2\left(\frac{1}{4n}\sum_{x_1, x_2 \in X} (x_1 - x_2)^2 \frac{1}{4n}\sum_{y_1, y_2 \in Y} (y_1 - y_2)^2 \right)^{\frac{-n+3}{2}} }}
\end{align}
\end{linenomath}

Canceling the fraction yields the result:
\begin{linenomath}
\begin{align}
   BC(p(\cdot|X), p(\cdot|Y) &= \left( 4 \frac{\sqrt{\sum_{x_1, x_2 \in X}(x_1 - x_2)^2 \sum_{y_1, y_2 \in Y}(y_1 - y_2)^2}}{\sum_{z_1, z_2 \in Z}(z_1 - z_2)^2} \right)^{\frac{n-3}{2}}\label{eq:gaus_quad_sum}
\end{align}
\end{linenomath}

The quadratic terms can be simplified individually:
\begin{linenomath}
\begin{align}
   \sum_{x_1, x_2 \in X}\left(x_1 - x_2\right)^2 = &\sum_{x_1\in X} \left(2n x_1^2 - 2\sum_{x_2\in X} x_1 x_2\right)\\
                                        = &2n \sum_{x_1\in X} x_1^2 - 2\sum_{x_1\in X} x_1 \sum_{x_2\in X} x_2\\
                                        = &2n \left(\sum_{x_1\in X} x_1^2 - n\left(\frac{1}{{n}}\sum_{x_1\in X} x_1\right) \left(\frac{1}{n}\sum_{x_1\in X} x_1\right)\right)\\
                               = &2n \left(n \mathbb{E} \left(X^2\right) - n\mathbb{E} \left(X\right)^2\right)\\
                                        = &2n^2 Var\left(X\right)
\end{align}
\end{linenomath}
Note, that we used the \textit{biased} estimator of the variance: $Var(X) = \frac{1}{n} \sum_{x \in X} (x - \mathbb{E}(X))^2$. Using the unbiased estimator would imply a factor of $\frac{n}{n-1}$. Since that is just a stylistic change, we decided to use this version to improve readability.
The simplification for the other quadratic terms work accordingly. Inserting into \autoref{eq:gaus_quad_sum} and further simplifications yield:

\begin{linenomath}
\begin{align}
   BC(p(\cdot|X), p(\cdot|Y) &= \left( 4 \frac{\sqrt{2n^2 Var(X) 2n^2 Var(Y)}}{2(2n)^2 Var(X \cup Y)} \right)^{\frac{n-3}{2}}\\
                                     &= \left(\frac{\sqrt{Var(X)Var(Y)}}{Var(X \cup Y)} \right)^{\frac{n-3}{2}}
\end{align}
\end{linenomath}
Since this is a very compact representation, we decided to display this version in the paper. We can however further transform the equation, to get a representation that has further advantages for numerical computations.

To do so, we use the following equality for the denominator:
\begin{linenomath}
\begin{align}\label{eq:simp_gauss_glob}
	Var(X \cup Y) &= \frac{1}{2n} \sum_{z_1\in X\cup Y} \left( z_1 - \frac{1}{2n}\sum_{z_2\in X\cup Y}  z_2\right)^2 \\
			      &= \frac{1}{2} \left( \frac{1}{n} \sum_{x\in X}\left( x - \frac{1}{2n}\sum_{z\in X\cup Y}  z\right)^2 + \frac{1}{n} \sum_{y\in Y}\left( y - \frac{1}{2n}\sum_{z\in X\cup Y}  z\right)^2 \right) \label{eq:var_sus}
\end{align}
\end{linenomath}

We can now analogously transform both summands like shown here for the first summand:
\begin{linenomath}
\begin{align}
	& \frac{1}{n} \sum_{x\in X}\left( x - \frac{1}{2n}\sum_{z\in X\cup Y} z\right)^2 \\
= & \frac{1}{n} \sum_{x\in X}\left( x - \frac{1}{n}\sum_{x_2\in X} x_2 + \frac{1}{n}\sum_{x_2\in X} x_2 - \frac{1}{2n}\sum_{z\in X\cup Y} z\right)^2 \\
      = & \frac{1}{n} \sum_{x\in X}\Bigg( \left( x - \frac{1}{n}\sum_{x_2\in X} x_2 \right)^2 + 2\left( x - \frac{1}{n}\sum_{x_2\in X} x_2 \right) \left(\frac{1}{n}\sum_{x_2\in X} x_2 - \frac{1}{2n}\sum_{z \in X\cup Y} z\right)  \nonumber\\
	& + \left(\frac{1}{n}\sum_{x_2\in X} x_2 - \frac{1}{2n}\sum_{z\in X\cup Y} z\right)^2 \Bigg)\\
      = & \frac{1}{n} \sum_{x\in X}\left( x - \frac{1}{n}\sum_{x_2\in X} x_2 \right)^2 + 2 \left( \frac{1}{n} \sum_{x\in X}  x - \frac{1}{n}\sum_{x_2\in X} x_2 \right) \left(\frac{1}{n}\sum_{x_2\in X} x_2 - \frac{1}{2n}\sum_{z \in X\cup Y} z\right)\nonumber\\
	& + \frac{1}{n} \sum_{x\in X} \left(\frac{1}{n}\sum_{x_2\in X} x_2 - \frac{1}{2n}\sum_{z \in X\cup Y} z\right)^2 \label{eq:sum_no_i} \\
      = & Var(X) + (\mathbb{E}(X) - \mathbb{E}(X \cup Y))^2
\end{align}
\end{linenomath}
where we used for the last equation, that the first factor of the second summand in \autoref{eq:sum_no_i} is 0 and that the inner summands of the third summand do not depend on $x$.
Next, we calculate
\begin{linenomath}
\begin{align}
	\mathbb{E}(X) - \mathbb{E}(X\cup Y) = \mathbb{E}(X) - \frac{\mathbb{E}(X) + \mathbb{E}(Y)}{2} = \frac{\mathbb{E}(X) - \mathbb{E}(Y)}{2}
\end{align}
\end{linenomath}
Applying this to both summands in \autoref{eq:var_sus} gives us
\begin{linenomath}
\begin{align}
	Var(X\cup Y) =& \frac{1}{2} \left(Var(X) + Var(Y) + \left(\frac{\mathbb{E}(X) - \mathbb{E}(Y)}{2}\right)^2 + \left(\frac{\mathbb{E}(X) - \mathbb{E}(Y)}{2}\right)^2 \right) \\
	=& \frac{Var(X) + Var(Y)}{2} + \left(\frac{\mathbb{E}(X) - \mathbb{E}(Y)}{2}\right)^2
\end{align}
\end{linenomath}
The whole equation for the BC distance then reads
\begin{linenomath}
\begin{align}
BC(p(\cdot|X), p(\cdot|Y) = \left(\frac{\sqrt{Var(X)Var(Y)}}{\frac{Var(X) + Var(Y)}{2} + \left(\frac{\mathbb{E}(X) - \mathbb{E}(Y)}{2}\right)^2
} \right)^{\frac{n-3}{2}}
\end{align}
\end{linenomath}
This formulation is preferable for numerical computations, since we do not have to calculate $Var(X\cup Y)$, which is time saving.

\end{document}